%% file: main.tex
\documentclass[11pt]{article}
\usepackage[final]{acl}

\usepackage{times}
\usepackage{amsmath,amssymb}
\usepackage{dsfont}
\usepackage{latexsym}

\usepackage[T1]{fontenc}

\usepackage[utf8]{inputenc}

\usepackage{microtype}

\usepackage{inconsolata}

\usepackage{graphicx}

\usepackage{booktabs}
\usepackage{multirow}

\usepackage{algorithm}
\usepackage{algpseudocode}

\usepackage{subcaption}

\usepackage{float}

\title{Beyond Similarity: Coverage-Aware Prompt Selection \\ for Time Series Forecasting with LLMs}

\author{
  Daeun Ji\thanks{\ \ Equal contribution.}\ \ \ \ Minkyoung Kim\footnotemark[1]\ \ \ \ Dongkuk Kim \\
  \textbf{Yohan Lee}\ \ \ \ \textbf{Beomsoo Kim}\thanks{\ \ Corresponding authors.}\ \ \ \ \textbf{Beakcheol Jang}\footnotemark[2] \\
  Graduate School of Information, Yonsei University, Seoul, Republic of Korea \\
  \texttt{jidaeun59@gmail.com}\ \ \ \ \texttt{minky@yonsei.ac.kr}\ \ \ \ \texttt{dongkuk@yonsei.ac.kr} \\
  \texttt{utopiamath@yonsei.ac.kr}\ \ \ \ \texttt{beomsoo@yonsei.ac.kr}\ \ \ \ \texttt{bjang@yonsei.ac.kr}
}

\newcommand{\stdv}[1]{{\tiny$\pm$#1}}

\makeatletter
\def\@fnsymbol#1{\ensuremath{\ifcase#1\or \dagger\or *\else\@ctrerr\fi}}
\makeatother

\begin{document}

\maketitle
\raggedbottom

\input{00_abstract}

\input{01_intro}

\section{Related Work}
\input{02_related}

\input{03_method}

\section{Experiments}
\input{04_experiments}

\input{05_discussion}

\input{07_conclusion}

\input{06_limitations}

\section*{Acknowledgments}
This work was supported by the Ministry of Education of the Republic of Korea and the National Research Foundation of Korea (NRF-2024S1A5C3A03046579).

The authors used an LLM assistant (Claude, Anthropic) to help revise the manuscript and take full responsibility for the content.

\bibliography{ref}

\appendix
\input{08_app_setup}
\input{09_app_mechanism}
\input{10_app_algorithm}
\input{11_app_longterm}
\input{12_app_shortterm}
\input{13_app_fewshot}
\input{14_app_sensitivity}
\input{12_app_significance}
\input{15_app_qualitative}

\end{document}

%% file: 00_abstract.tex
\begin{abstract}
Similarity-based retrieval is the dominant rule for conditioning large language models (LLMs) in in-context learning, retrieval-augmented generation, and prompt-based time series forecasting. The rule concentrates on near-duplicate candidates, an issue that has motivated diversity-aware retrieval but remains unexamined in other retrieval-conditioned pipelines. We study this issue using prompt-based time series forecasting as a test bed, where a learned prompt pool is retrieved by similarity. Dominant methods in this setting retrieve top-$K$ entries by cosine similarity without redundancy control, producing a bias toward dominant temporal patterns while overlooking rare but informative events. We propose CASP-LLM, a coverage-aware semantic prompting framework that addresses this prompt selection bias by combining usage-tracking and saturating-gate techniques into a coverage regularizer that adds no learnable parameters. On six long-term benchmarks and the M4 short-term benchmark, CASP-LLM matches or improves on similarity-based LLM forecasters on most dataset-horizon settings, with the exceptions of Electricity, M4-Monthly, and the few-shot long-horizon setting. A controlled study locates the failure mode at the cross-batch usage level rather than per-retrieval redundancy: within-retrieval diversification such as MMR does not help, whereas regularizing anchor usage across training does.
\end{abstract}

%% file: 01_intro.tex
\vspace{10pt}

\section{Introduction}

Conditioning large language models on retrieved candidates from a learned pool is a recurring design pattern across applications such as in-context learning and retrieval-augmented generation. The dominant rule, top-$K$ cosine similarity, has a well-documented redundancy issue: selected candidates collapse onto near-duplicates and fail to span the representational diversity of the pool~\cite{ye2023complementary, gupta2023coverage, liu2024se2}. Existing remedies trade off relevance against inter-candidate diversity, including maximal marginal relevance~\cite{carbonell1998use}, determinantal point processes~\cite{kulesza2012determinantal}, and submodular coverage~\cite{lin2011submodular}.

\begin{figure}[t!]
\centering
\includegraphics[width=\columnwidth]{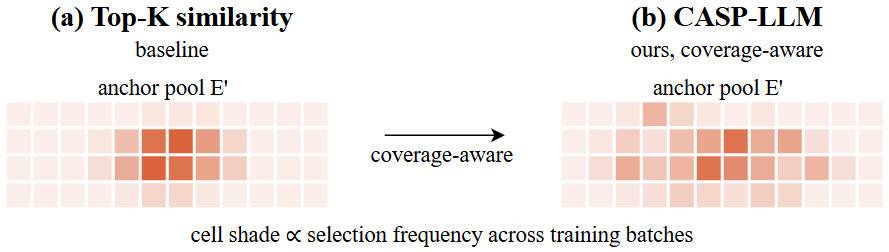}
\caption{Selection bias in similarity-based retrieval. (a) Top-$K$ cosine similarity concentrates selection on a narrow subset of semantic anchors across training batches. (b) CASP-LLM redistributes anchor usage during training through a coverage penalty while preserving similarity-based relevance. Cell shade is proportional to selection frequency accumulated over training batches.}
\label{intro}
\end{figure}

While these remedies are established in in-context example selection and retrieval-augmented generation, the same issue remains largely unexamined in other retrieval-conditioned LLM pipelines. We study it in large language model (LLM)-based time series forecasting, where a small, learned, explicitly retrieved prompt pool makes the selection mechanism and its effects directly measurable. Recent methods such as TEMPO~\cite{cao2023tempo} and $\text{S}^2$IP-LLM~\cite{pan2024s2ip} construct a learned pool of soft prompts or semantic anchors derived from pretrained LLM embeddings, then retrieve top-$K$ entries by cosine similarity to the time-series query. These methods inherit the same redundancy pathology, producing a bias toward dominant temporal patterns while overlooking rare but informative events. We call this failure mode \emph{prompt selection bias}. As shown in Figure~\ref{intro}, similarity-driven approaches concentrate selection on a narrow subset of the anchor pool, analogous to the over- and under-translation that motivated coverage mechanisms in neural machine translation~\cite{tu2016modeling}. The retrieved set conveys a single dominant temporal mode rather than complementary modes, and the cost of that narrowing falls on inputs that depart from the dominant mode, which is where the gains of the coverage term concentrate (Appendix~\ref{app:significance:strata}).

We propose CASP-LLM, a coverage-aware semantic prompting framework that addresses prompt selection bias by maintaining a per-anchor usage statistic and gating the optimization signal through a saturating $\min$ operator. The statistic is an exponential moving average over recent batches, as in running-statistic tracking in deep learning~\cite{tarvainen2017meanteacher}, and connects to usage balancing in mixture-of-experts routing~\cite{fedus2022switch}. The resulting regularizer is lightweight, adds no learnable parameters, and integrates with the training objective rather than acting only at inference time. The design connects to coverage in neural sequence generation~\cite{tu2016modeling, mi2016coverage, see2017get}, extending the principle of penalizing repeated attention from token-level decoding to retrieval-pool-level usage.

Our contributions are as follows:
\begin{itemize}
\item We identify \emph{prompt selection bias} as a failure mode of similarity-based retrieval-conditioning in LLM-based time series forecasting.
\item We instantiate this principle as a coverage-aware retrieval mechanism that combines usage tracking and a saturating gate, with a stop-gradient safeguard, into a regularizer that adds no learnable parameters.
\item We empirically evaluate CASP-LLM across multivariate datasets and forecasting horizons, where it matches or improves on strong similarity-based baselines, and we localize the failure mode to cross-batch usage rather than within-retrieval redundancy.
\end{itemize}

Empirically, CASP-LLM attains the best or second-best MSE on most dataset-horizon settings among similarity-based LLM forecasters across six long-term benchmarks and remains competitive on the M4 short-term benchmark.
An anchor replacement ablation shows that the retrieved anchors carry task-relevant structure beyond embedding-space proximity, and a coverage-strength sweep places the useful setting at a moderate weight rather than the largest one.

%% file: 02_related.tex
\subsection{Diversity-Aware Retrieval}
\label{sec:related:retrieval}
The limitations of similarity-based retrieval are well documented in in-context example selection and retrieval-augmented generation. KATE~\cite{liu2022kate} established the cosine-similarity baseline, EPR~\cite{rubin2022epr} learned a retriever from the inference model's likelihood of the gold output, and~\citet{lu2022fantastically} showed sensitivity to demonstration order. Since these methods score candidates independently, subsequent work introduced diversity-aware criteria: CEIL~\cite{ye2023complementary} formulates exemplar selection as conditional-DPP subset selection,~\citet{su2023votek} diversify the annotation pool through graph-based selection, Se$^2$~\cite{liu2024se2} reframes selection as sequential decision-making with diversity refinement, and~\citet{gupta2023coverage} argue that demonstrations should \emph{cover} test-instance substructures via set-level coverage objectives based on BERTScore recall. A parallel line formalizes selection as submodular optimization~\cite{qian2024subsa, nanda2025insquad, zheng2025subcp}, enforcing diversity or quality-diversity trade-offs over the selected subset. These methods all operate over discrete labeled exemplar pools and select per query, whereas our setting concerns a continuous semantic anchor pool and a usage signal accumulated across training.

Retrieval-augmented generation~\cite{lewis2020retrieval} faces the same issue with dense retrievers such as DPR~\cite{karpukhin2020dpr}, motivating diversity-aware re-ranking through maximal marginal relevance~\cite{carbonell1998use}, determinantal point processes~\cite{kulesza2012determinantal}, and submodular subset selection~\cite{lin2011submodular}. Our coverage mechanism is closest to these saturating-coverage formulations~\cite{lin2011submodular, carbonell1998use}, but with the saturation budget accumulating across batches rather than within a single call, and it further relates to coverage in neural sequence generation~\cite{tu2016modeling, mi2016coverage, see2017get}. As we discuss next, this diversity-aware perspective has not been transported to the retrieval-conditioned setting of LLM-based time series forecasting.

\subsection{LLM-Based Time Series Forecasting}
\label{sec:related:llm-ts}
Time series forecasting has progressed from statistical and recurrent models~\cite{hochreiter1997longLSTM, chung2014GRU} to transformer-based architectures with decomposition and patching designs~\cite{wu2021autoformer, zhou2022fedformer, wu2023timesnet, nie2023patchtst, liu2023itransformer}. These methods process input sequences uniformly and rely on the inductive biases of the architecture itself, without an explicit mechanism for selectively reusing historical context or incorporating semantic priors from large pretrained models. Recent work adapts pretrained LLMs as forecasting backbones, freezing or partially fine-tuning the LLM and treating numerical patches as tokens or aligning them with word-embedding spaces~\cite{zhou2023one, jin2024time, chang2025llm4ts, xue2023promptcast, liu2025calf, liu2025timecma}, with the broader goal of transferring the linguistic inductive bias of pretrained LMs to non-textual sequence modeling.

A subset of these methods constructs a learned pool of prompts or semantic anchors and retrieves entries by similarity, which is the direct comparison setting for our work. TEMPO~\cite{cao2023tempo} maintains a bank of learnable prompts inspired by the L2P prompt-pool design from continual learning~\cite{wang2022l2p} and retrieves the top-$K$ entries by cosine similarity; $\text{S}^2$IP-LLM~\cite{pan2024s2ip} derives semantic anchors from pretrained GPT-2 word embeddings and selects the top-$K$ most similar to the time-series representation as prefix prompts. These methods inherit the similarity-based selection rule discussed in Section~\ref{sec:related:retrieval} without any redundancy control. They differ in where such a control can be attached: $\text{S}^2$IP-LLM draws its anchors from a fixed embedding table, so usage is well defined over a stable anchor index and the top-$K$ rule is preserved unchanged, whereas TEMPO ties its prompts to a trend, seasonal, and residual decomposition, which places selection at a different point in the pipeline. We instantiate the coverage term on the anchor-based design; Appendix~\ref{app:setup:baselines} states what a matched comparison on the prompt-bank design would involve. A concurrent line of retrieval-augmented forecasters~\cite{yang2025timerag, zhang2024timeraf, ning2025tsrag, han2025raft} retrieves reference sequences from an external store by raw similarity or DTW; these target what to retrieve from a corpus rather than how to control redundancy among selected entries, leaving the selection bias we study unaddressed. CASP-LLM keeps this similarity-based skeleton and the top-$K$ rule unchanged, and instead augments training with a coverage-aware regularizer that accumulates anchor usage across batches~(Section~\ref{sec:method:coverage}).

%% file: 03_method.tex
\section{Methodology}
\label{sec:method}

\begin{figure*}[h!]
\centering
    \includegraphics[width=\textwidth]{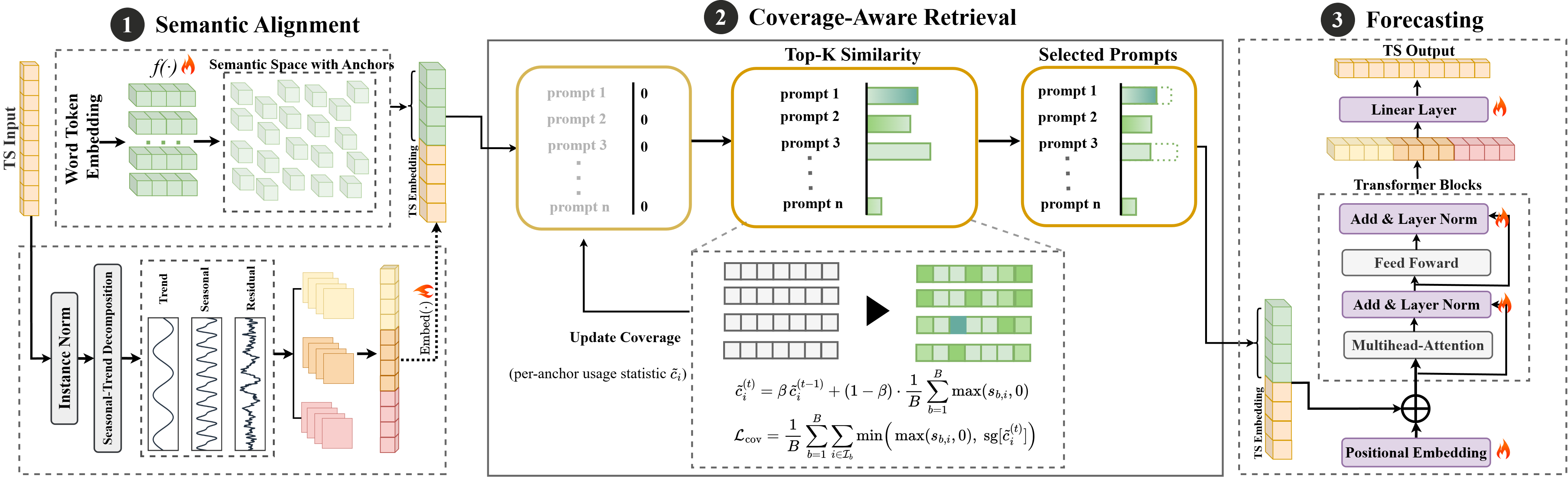}
    \caption{Overall architecture of the CASP-LLM framework. The framework consists of three main stages: (1) Semantic Alignment of multivariate time-series patches with the LLM semantic space, (2) Coverage-Aware Retrieval that balances semantic similarity with anchor usage across training, and (3) Prompt-Augmented Forecasting via the GPT decoder.}
    \label{fig:over}
\end{figure*}

Figure~\ref{fig:over} provides an overview of the framework, whose central contribution is the coverage-aware retrieval mechanism (Section~\ref{sec:method:coverage}). The surrounding components, including time-series alignment, semantic anchor matching, and prompt-augmented decoding, follow established designs in the LLM-for-TS literature~\cite{nie2023patchtst, pan2024s2ip, zhou2023one}. Within this skeleton, the coverage-aware mechanism augments the similarity-based selection rule with a usage-penalized training objective that addresses the prompt selection bias described in the introduction, while leaving the top-$K$ retrieval rule itself unchanged.

\subsection{Problem Definition}

Given a multivariate time series $\mathbf{X} \in \mathbb{R}^{B \times L \times M}$ with batch size $B$, input length $L$, and $M$ variables, we predict future values $\hat{\mathbf{Y}} \in \mathbb{R}^{B \times P \times M}$ over the next $P$ steps as a prompt-augmented sequence-to-sequence problem.

\subsection{Semantic Alignment}

This stage maps multivariate time-series inputs and the pretrained word-token vocabulary into a shared semantic space, producing the query and anchor representations consumed by the retrieval mechanism in Section~\ref{sec:method:coverage}.

\paragraph{Time-series branch}
For each batch sample $b$, the input $\mathbf{X}_b \in \mathbb{R}^{L \times M}$ is divided along the temporal axis into $N_p$ overlapping patches of length $P_{\text{patch}}$ following PatchTST~\cite{nie2023patchtst}, yielding $\mathbf{x}_{b,n} \in \mathbb{R}^{P_{\text{patch}} \times M}$; the $M$ variables are processed independently, so the batch index $b$ below ranges over sample-variable pairs. Each patch is decomposed additively into trend, seasonal, and residual components $\mathbf{x}_{b,n}^T, \mathbf{x}_{b,n}^S, \mathbf{x}_{b,n}^R$ by classical seasonal-trend decomposition, following the tokenization design of $\text{S}^2$IP-LLM~\cite{pan2024s2ip}, which are concatenated along the channel dimension and flattened into a unified patch vector $\mathbf{x}'_{b,n} \in \mathbb{R}^{D}$ with $D = 3P_{\text{patch}}$, preserving disentangled structural information. Each patch vector is mapped into the LLM embedding space by a learnable linear projection $\mathbf{e}_{b,n} = \mathrm{Embed}(\mathbf{x}'_{b,n}) \in \mathbb{R}^{d}$, where $d$ is the LLM embedding dimension. Patch embeddings are then averaged across $n$ to obtain a sample-level query representation $\tilde{\mathbf{e}}_b = \frac{1}{N_p}\sum_{n=1}^{N_p} \mathbf{e}_{b,n}$.

\paragraph{Anchor branch}
Prior LLM-for-TS approaches~\cite{cao2023tempo} typically formulate soft prompts as randomly initialized continuous vectors optimized through the forecasting loss, without explicitly leveraging the semantic space acquired during LLM pre-training. We instead derive prompts from the pretrained semantic space. Let $\mathbf{E} \in \mathbb{R}^{V \times d}$ denote the input word-token embedding table of the pretrained LLM, with vocabulary size $V$ and embedding dimension $d$. To avoid the cost of using $\mathbf{E}$ directly given large $V$, and following the semantic anchor design of $\text{S}^2$IP-LLM~\cite{pan2024s2ip}, we apply a learnable linear projection $f(\cdot)$ to extract a small set of representative vectors:
\begin{equation}
\mathbf{E}' = f(\mathbf{E}) \in \mathbb{R}^{V' \times d},
\end{equation}
where $V' \ll V$. Each row $\mathbf{p}_i \in \mathbb{R}^{d}$ for $i = 1, \dots, V'$ is a \emph{semantic anchor}. Since $f(\cdot)$ is linear, each anchor $\mathbf{p}_i$ is a learned linear combination of pretrained word embeddings and does not correspond to any specific vocabulary token; $f$ extracts $V'$ representative directions from the pretrained semantic manifold rather than selecting individual words.

\subsection{Coverage-Aware Retrieval}
\label{sec:method:coverage}

We align the query $\tilde{\mathbf{e}}_b$ with the anchors via cosine similarity:
\begin{equation}
s_{b,i} = \hat{\mathbf{e}}_b \cdot \tilde{\mathbf{p}}_i,
\end{equation}
where $\hat{\mathbf{e}}_b = \tilde{\mathbf{e}}_b / \|\tilde{\mathbf{e}}_b\|_2$ and $\tilde{\mathbf{p}}_i = \mathbf{p}_i / \|\mathbf{p}_i\|_2$ are the L2-normalized query and anchor, so that $s_{b,i} \in [-1, 1]$. Semantically aligned anchors typically yield non-negative values in practice. Naively retrieving the top-$K$ anchors by $s_{b,i}$ leads to selection collapse, the prompt selection bias of the introduction: prompt selection concentrates on a small set of highly similar anchors, and this bias accumulates across training batches as the embedding space reorganizes to favor those same anchors. Existing remedies for similarity-based redundancy in retrieval, including maximal marginal relevance~\cite{carbonell1998use}, determinantal point processes~\cite{kulesza2012determinantal}, and submodular coverage~\cite{lin2011submodular}, operate within a single retrieval call by re-ranking or selecting among candidates based on pairwise diversity, and therefore do not directly counter this cross-batch accumulation. We instead propose a coverage-aware regularizer that tracks per-anchor usage across batches and gates the optimization signal accordingly, addressing the bias at the training-dynamics level rather than the per-call selection level.

We introduce a coverage statistic that tracks the recent usage of each anchor and a loss term that gates the optimization signal on this statistic. The construction combines two techniques: exponential moving average (EMA) usage tracking of running statistics in deep learning~\cite{tarvainen2017meanteacher}, related to usage-balancing objectives in mixture-of-experts routing~\cite{fedus2022switch}, and a saturating gate that bounds the regularization signal, with a stop-gradient that decouples the statistic from the optimizer.

\paragraph{Anchor usage tracker}
Let $\tilde{c}_i^{(t)} \in \mathbb{R}$ denote the coverage statistic for anchor $i$ at training step $t$. We update $\tilde{c}_i^{(t)}$ via an exponential moving average over the per-batch mean similarity:
\begin{equation}
\label{eq:cov_update}
\tilde{c}_i^{(t)} = \beta \, \tilde{c}_i^{(t-1)} + (1 - \beta) \, \bar{s}_i^{(t)}
\end{equation}
\begin{equation}
\bar{s}_i^{(t)} = \frac{1}{B} \sum_{b=1}^{B} \max(s_{b,i}^{(t)}, 0),
\end{equation}
where $\beta \in (0, 1)$ is a decay factor. The clipping at zero ensures that $\tilde{c}_i^{(t)}$ lives on the same scale as the similarity score $s_{b,i}$, which is essential for the saturating gate below. The update is applied only during training; during validation and inference, $\tilde{c}_i$ is held fixed.

\paragraph{Top-$K$ selection}
The top-$K$ anchors are selected according to the similarity score $s_{b,i}$:
\begin{equation}
\mathcal{I}_b = \text{TopK}(\mathbf{s}_b, K).
\end{equation}
This retrieval rule is preserved in CASP-LLM; the coverage mechanism instead operates through the loss function, biasing the embedding geometry during training so that under-utilized anchors become more accessible to selection.

\paragraph{Composite objective}
Model optimization jointly considers forecasting accuracy, semantic alignment, and coverage through a composite objective:
\begin{equation}
\mathcal{L}_{\text{total}} = \mathcal{L}_{\text{forecast}} + \lambda_{\text{sim}} \mathcal{L}_{\text{sim}} + \lambda_{\text{cov}} \mathcal{L}_{\text{cov}}.
\end{equation}
The forecasting loss $\mathcal{L}_{\text{forecast}}$ measures prediction error against the ground truth. The similarity loss penalizes the gap between the selected anchors and the query:
\begin{equation}
\mathcal{L}_{\text{sim}} = \frac{1}{B} \sum_{b=1}^{B} \sum_{i \in \mathcal{I}_b} (1 - s_{b,i}).
\end{equation}
The coverage loss is defined through a saturating $\min$ gate:
\begin{equation}
\label{eq:cov_loss}
\mathcal{L}_{\text{cov}} = \frac{1}{B} \sum_{b=1}^{B} \sum_{i=1}^{V'} \min\!\bigl(\max(s_{b,i}, 0), \, \mathrm{sg}[\tilde{c}_i^{(t)}]\bigr),
\end{equation}
where $\mathrm{sg}[\cdot]$ denotes stop-gradient.

The $\min$ operator implements a saturating gate on the coverage penalty. For an anchor whose accumulated usage $\tilde{c}_i$ is high, the gate returns $s_{b,i}$ for every query with $s_{b,i} < \tilde{c}_i$, so the penalty lowers that anchor's similarity over a wide range of queries. For an anchor with $\tilde{c}_i$ near zero, the gate saturates at $\tilde{c}_i$ for almost every query and the penalty is inactive, leaving the anchor untouched. Coverage pressure therefore falls on the anchors that dominate retrieval and, by lowering their similarity, makes less-used anchors competitive under the unchanged top-$K$ rule. The stop-gradient on $\tilde{c}_i$ prevents the optimizer from gaming the regularizer by collapsing $s_{b,i}$ globally to keep $\tilde{c}_i$ small. Appendix~\ref{app:mechanism} gives the gradient in closed form and discusses each design choice.

\subsection{Prompt-Augmented Forecasting}

We use GPT-2~\cite{radford2019language} as the decoder backbone. For each sample, the top-$K$ selected anchor embeddings are concatenated as a prefix to the time-series patch embeddings, serving as contextual control tokens that condition decoding without being prediction targets. The sequence is processed with causal attention, and forecasting is performed by projecting the hidden states corresponding to the prediction horizon through a linear output layer. Following the partial fine-tuning recipe adopted by OFA~\cite{zhou2023one} and $\text{S}^2$IP-LLM~\cite{pan2024s2ip}, we freeze the multi-head attention and feed-forward layers and fine-tune only the positional embedding and layer normalization, which has been reported to perform comparably to or better than full fine-tuning in this setting.

\paragraph{Training and Inference}
During training the model is optimized with $\mathcal{L}_{\text{total}}$ and the coverage vector $\tilde{\mathbf{c}}$ is updated via Equation~(\ref{eq:cov_update}) at each step; at validation and test time only the forecasting loss is evaluated.

%% file: 04_experiments.tex
\subsection{Long- and Short-term Forecasting}
\label{sec:exp:longterm}

We evaluate CASP-LLM on six multivariate long-term forecasting benchmarks (Weather, ETTh1/h2, ETTm1/m2, Electricity) with input length 512, forecasting horizons $\{96, 192, 336, 720\}$, and metrics MSE and MAE, averaged over three runs. We follow the experimental configurations of \citet{wu2023timesnet} for all baselines and our model's basic modules; baselines and additional implementation details appear in Appendix~\ref{app:setup}. Table~\ref{tab:longterm} presents long-term forecasting results; CASP-LLM entries are means over three seeds with sample standard deviations. CASP-LLM attains the best or second-best MSE in 18 of 24 dataset--horizon cells (best in 14) and the best or second-best MAE in 17 of 24; the exceptions concentrate on Electricity and the hourly ETT benchmarks.\footnote{Cells outside the top two in MSE: Electricity at 336 and 720, ETTh1 at 96, 192, and 720, and ETTh2 at 720. On the horizon-averaged Electricity comparison, iTransformer leads with MSE 0.1640 and MAE 0.2584 against 0.1702 and 0.2674 for CASP-LLM (Table~\ref{tab:full_results_avg}).} This behavior stems from two architectural characteristics of CASP-LLM. First, the model projects time-series patch representations into a pretrained semantic embedding space, enabling explicit alignment between numerical inputs and semantic anchors. Second, prompt selection is regularized through a coverage-aware mechanism that jointly considers semantic similarity and historical anchor usage, mitigating the prompt selection bias inherent in pure similarity-based retrieval. We additionally report short-term forecasting results on the M4 benchmark~\cite{makridakis2020m4} in Appendix~\ref{app:shortterm}, where CASP-LLM remains competitive in average sMAPE and MASE with LLM-based and non-LLM baselines.

Our claims about the coverage term rest on the paired comparison against the $\lambda_{\text{cov}}=0$ configuration, which holds code, data, seeds, and tuning identical, rather than on rankings against external baselines. ETTh1-96, the one cell where the eight-seed paired test reaches significance (Appendix~\ref{app:significance}), serves as the primary cell for the mechanism analyses in Section~\ref{sec:exp:ablation}.

\begin{table*}[h!]
\centering
\caption{Long-term forecasting results for $\{96, 192, 336, 720\}$ horizons across six multivariate benchmarks. CASP-LLM entries are three-seed means$\pm$std. Lower is better; best in bold, second-best underlined.}
\label{tab:longterm}
\scriptsize
\setlength{\tabcolsep}{3pt}
\resizebox{\textwidth}{!}{%
\begin{tabular}{@{}l|c||cc|cc|cc|cc|cc|cc|cc|cc@{}}
\toprule
& & \multicolumn{2}{c|}{\textbf{Ours}} & \multicolumn{2}{c|}{\textbf{Time-LLM}} & \multicolumn{2}{c|}{\textbf{S$^2$IP-LLM}} & \multicolumn{2}{c|}{\textbf{CALF}} & \multicolumn{2}{c|}{\textbf{OFA}} & \multicolumn{2}{c|}{\textbf{iTransformer}} & \multicolumn{2}{c|}{\textbf{PatchTST}} & \multicolumn{2}{c}{\textbf{DLinear}} \\
\cline{3-18}
\textbf{Dataset} & \textbf{H} & MSE & MAE & MSE & MAE & MSE & MAE & MSE & MAE & MSE & MAE & MSE & MAE & MSE & MAE & MSE & MAE \\
\midrule
\multirow{4}{*}{Weather} & 96 & \textbf{0.1483}\stdv{0.0013} & \textbf{0.2007}\stdv{0.0019} & 0.1663 & 0.2200 & 0.1506 & \underline{0.2011} & 0.1571 & 0.2071 & \underline{0.1498} & 0.2020 & 0.1708 & 0.2212 & 0.1551 & 0.2123 & 0.1704 & 0.2298 \\
 & 192 & \textbf{0.1935}\stdv{0.0007} & 0.2445\stdv{0.0022} & 0.2024 & 0.2501 & \underline{0.1941} & \textbf{0.2426} & 0.2036 & 0.2494 & 0.1944 & \underline{0.2429} & 0.2118 & 0.2564 & 0.2122 & 0.2636 & 0.2140 & 0.2719 \\
 & 336 & \textbf{0.2445}\stdv{0.0025} & \underline{0.2821}\stdv{0.0036} & 0.2500 & 0.2855 & \underline{0.2450} & \textbf{0.2810} & 0.2523 & 0.2873 & 0.2482 & 0.2848 & 0.2604 & 0.2933 & 0.2486 & 0.2857 & 0.2596 & 0.3112 \\
 & 720 & \textbf{0.3181}\stdv{0.0026} & \underline{0.3345}\stdv{0.0041} & 0.3197 & \textbf{0.3338} & 0.3237 & 0.3383 & 0.3322 & 0.3437 & 0.3216 & 0.3369 & 0.3341 & 0.3391 & \underline{0.3192} & 0.3378 & 0.3225 & 0.3634 \\
\midrule
\multirow{4}{*}{ETTh1} & 96 & 0.3714\stdv{0.0036} & \underline{0.4003}\stdv{0.0009} & 0.4136 & 0.4292 & \underline{0.3711} & 0.4007 & 0.3863 & 0.4135 & 0.3833 & 0.4049 & 0.3981 & 0.4227 & 0.3834 & 0.4122 & \textbf{0.3682} & \textbf{0.3981} \\
 & 192 & 0.4161\stdv{0.0139} & 0.4334\stdv{0.0117} & 0.4240 & 0.4337 & \underline{0.4026} & \underline{0.4217} & 0.4176 & 0.4323 & 0.4201 & 0.4273 & 0.4306 & 0.4450 & 0.4696 & 0.4675 & \textbf{0.4025} & \textbf{0.4205} \\
 & 336 & \underline{0.4263}\stdv{0.0183} & \underline{0.4423}\stdv{0.0127} & 0.4503 & 0.4583 & \textbf{0.4229} & 0.4442 & 0.4315 & 0.4432 & 0.4356 & \textbf{0.4375} & 0.4429 & 0.4560 & 0.4712 & 0.4687 & 0.4311 & 0.4425 \\
 & 720 & 0.5057\stdv{0.0296} & 0.4957\stdv{0.0164} & 0.4757 & 0.4854 & \textbf{0.4649} & \textbf{0.4719} & 0.4769 & 0.4844 & \underline{0.4670} & \underline{0.4726} & 0.5224 & 0.5212 & 0.5158 & 0.5089 & 0.4718 & 0.4932 \\
\midrule
\multirow{4}{*}{ETTh2} & 96 & \textbf{0.2846}\stdv{0.0040} & \textbf{0.3456}\stdv{0.0051} & 0.3068 & 0.3627 & \underline{0.2856} & \underline{0.3456} & 0.2864 & 0.3462 & 0.2912 & 0.3458 & 0.3038 & 0.3594 & 0.3154 & 0.3715 & 0.3025 & 0.3676 \\
 & 192 & \textbf{0.3474}\stdv{0.0011} & \textbf{0.3887}\stdv{0.0033} & 0.3700 & 0.3979 & \underline{0.3597} & 0.3977 & 0.3734 & 0.4001 & 0.3656 & \underline{0.3950} & 0.3755 & 0.4049 & 0.3947 & 0.4234 & 0.3941 & 0.4261 \\
 & 336 & \textbf{0.3607}\stdv{0.0051} & \textbf{0.4080}\stdv{0.0037} & 0.3804 & 0.4143 & \underline{0.3662} & \underline{0.4121} & 0.3867 & 0.4158 & 0.4032 & 0.4284 & 0.4182 & 0.4345 & 0.4166 & 0.4415 & 0.4900 & 0.4895 \\
 & 720 & 0.4203\stdv{0.0066} & 0.4557\stdv{0.0055} & \textbf{0.4173} & \underline{0.4473} & 0.4381 & 0.4654 & \underline{0.4199} & \textbf{0.4458} & 0.4369 & 0.4567 & 0.4410 & 0.4628 & 0.4876 & 0.4855 & 0.8112 & 0.6381 \\
\midrule
\multirow{4}{*}{ETTm1} & 96 & \underline{0.2965}\stdv{0.0044} & 0.3529\stdv{0.0016} & 0.3303 & 0.3685 & \textbf{0.2928} & \underline{0.3496} & 0.3006 & 0.3503 & 0.2996 & 0.3547 & 0.3104 & 0.3639 & 0.3135 & 0.3625 & 0.3037 & \textbf{0.3471} \\
 & 192 & \textbf{0.3285}\stdv{0.0023} & \underline{0.3721}\stdv{0.0024} & 0.3581 & 0.3843 & \underline{0.3324} & 0.3722 & 0.3370 & 0.3761 & 0.3395 & 0.3802 & 0.3481 & 0.3874 & 0.3640 & 0.3981 & 0.3355 & \textbf{0.3662} \\
 & 336 & \textbf{0.3557}\stdv{0.0026} & \textbf{0.3874}\stdv{0.0003} & 0.3869 & 0.3993 & \underline{0.3625} & 0.3919 & 0.3682 & 0.3946 & 0.3681 & 0.3932 & 0.3797 & 0.4045 & 0.3918 & 0.4135 & 0.3714 & \underline{0.3909} \\
 & 720 & \textbf{0.4113}\stdv{0.0033} & \textbf{0.4167}\stdv{0.0008} & 0.4407 & 0.4344 & \underline{0.4160} & \underline{0.4203} & 0.4262 & 0.4268 & 0.4312 & 0.4315 & 0.4375 & 0.4396 & 0.4388 & 0.4449 & 0.4236 & 0.4223 \\
\midrule
\multirow{4}{*}{ETTm2} & 96 & \textbf{0.1651}\stdv{0.0024} & \underline{0.2577}\stdv{0.0019} & 0.1809 & 0.2756 & \underline{0.1665} & \textbf{0.2563} & 0.1721 & 0.2615 & 0.1728 & 0.2662 & 0.1801 & 0.2718 & 0.1768 & 0.2693 & 0.1676 & 0.2633 \\
 & 192 & \underline{0.2255}\stdv{0.0040} & \underline{0.3015}\stdv{0.0027} & 0.2310 & 0.3069 & \textbf{0.2247} & 0.3040 & 0.2294 & \textbf{0.2993} & 0.2334 & 0.3074 & 0.2421 & 0.3132 & 0.2414 & 0.3107 & 0.2257 & 0.3056 \\
 & 336 & \textbf{0.2760}\stdv{0.0046} & \underline{0.3352}\stdv{0.0046} & \underline{0.2795} & 0.3384 & 0.2862 & 0.3426 & 0.2829 & \textbf{0.3337} & 0.2868 & 0.3445 & 0.2920 & 0.3452 & 0.2928 & 0.3460 & 0.2987 & 0.3618 \\
 & 720 & \textbf{0.3549}\stdv{0.0027} & \textbf{0.3864}\stdv{0.0005} & 0.3682 & 0.3882 & \underline{0.3622} & 0.3916 & 0.3669 & \underline{0.3867} & 0.3721 & 0.3969 & 0.3728 & 0.3944 & 0.4079 & 0.4152 & 0.4150 & 0.4380 \\
\midrule
\multirow{4}{*}{Electricity} & 96 & \underline{0.1342}\stdv{0.0006} & \underline{0.2320}\stdv{0.0007} & 0.1463 & 0.2502 & 0.1378 & 0.2376 & 0.1373 & 0.2323 & 0.1373 & 0.2367 & \textbf{0.1339} & \textbf{0.2294} & 0.1369 & 0.2408 & 0.1406 & 0.2410 \\
 & 192 & \textbf{0.1511}\stdv{0.0002} & \textbf{0.2472}\stdv{0.0014} & 0.1587 & 0.2639 & 0.1567 & 0.2557 & 0.1571 & 0.2528 & 0.1534 & \underline{0.2506} & 0.1566 & 0.2509 & \underline{0.1533} & 0.2548 & 0.1544 & 0.2542 \\
 & 336 & 0.1736\stdv{0.0055} & 0.2732\stdv{0.0104} & 0.1737 & 0.2720 & 0.1710 & 0.2693 & \textbf{0.1677} & \textbf{0.2648} & \underline{0.1686} & 0.2668 & 0.1691 & \underline{0.2652} & 0.1691 & 0.2699 & 0.1693 & 0.2709 \\
 & 720 & 0.2220\stdv{0.0062} & 0.3170\stdv{0.0078} & 0.2089 & 0.3017 & 0.2094 & 0.3013 & \underline{0.2036} & \underline{0.2946} & 0.2061 & 0.2979 & \textbf{0.1963} & \textbf{0.2882} & 0.2058 & 0.3001 & 0.2040 & 0.3037 \\
\bottomrule
\end{tabular}}
\end{table*}

\subsection{Ablation and Structural Sensitivity Analysis}
\label{sec:exp:ablation}
We analyze how coverage-aware selection affects anchor usage and forecasting performance, validate the contributions of prompt alignment and decomposition, and verify that retrieved anchors carry task-relevant structure. Additional sensitivity analyses are reported in Appendix~\ref{app:sensitivity}.

\begin{table}[h!]
\centering
\caption{Coverage weight sweep on the primary cells. Entropy is hard-assignment usage entropy on the test split, with floor $\ln K$: $2.079$ for ETTh1 ($K=8$) and $1.386$ for ETTm1 ($K=4$). Cosine is the mean pairwise cosine among selected anchor keys.}
\scriptsize
\setlength{\tabcolsep}{3pt}
\resizebox{\columnwidth}{!}{%
\begin{tabular}{c|c|l||c|c|c|c}
\toprule
 &  & $\lambda_{\text{cov}}$ & 0 & 0.01 & 0.1 & 0.5 \\
\midrule
\multirow{6}{*}{ETTh1} & \multirow{3}{*}{96} & Entropy & $2.0794$ & $2.0794$ & $2.0991$\stdv{0.0341} & $2.2535$\stdv{0.1692} \\
  & & Cosine & $0.9971$\stdv{0.0017} & $0.9985$\stdv{0.0005} & $0.3657$\stdv{0.5483} & $0.1512$\stdv{0.0543} \\
  & & MSE & $0.3752$\stdv{0.0067} & $0.3746$\stdv{0.0074} & $\mathbf{0.3714}$\stdv{0.0036} & $0.3748$\stdv{0.0028} \\
\cline{2-7}
  & \multirow{3}{*}{336} & Entropy & $2.0794$ & $2.0794$ & $2.0794$ & $2.0794$ \\
  & & Cosine & $0.9730$\stdv{0.0291} & $0.9834$\stdv{0.0098} & $0.9293$\stdv{0.1116} & $0.9923$\stdv{0.0018} \\
  & & MSE & $0.4451$\stdv{0.0025} & $0.4331$\stdv{0.0156} & $\mathbf{0.4263}$\stdv{0.0183} & $0.4438$\stdv{0.0041} \\
\midrule
\multirow{6}{*}{ETTm1} & \multirow{3}{*}{96} & Entropy & $1.3863$ & $1.3863$ & $1.5003$\stdv{0.1975} & $1.3863$ \\
  & & Cosine & $0.9992$\stdv{0.0006} & $0.9985$\stdv{0.0019} & $0.4966$\stdv{0.5044} & $0.8296$\stdv{0.2810} \\
  & & MSE & $0.2937$\stdv{0.0032} & $0.2936$\stdv{0.0024} & $0.2920$\stdv{0.0017} & $\mathbf{0.2915}$\stdv{0.0005} \\
\cline{2-7}
  & \multirow{3}{*}{336} & Entropy & $1.3863$ & $1.3863$ & $1.3863$ & $1.3863$ \\
  & & Cosine & $0.9994$\stdv{0.0003} & $0.9993$\stdv{0.0004} & $0.9982$\stdv{0.0014} & $0.6846$\stdv{0.2665} \\
  & & MSE & $0.3577$\stdv{0.0009} & $\mathbf{0.3557}$\stdv{0.0026} & $0.3589$\stdv{0.0050} & $0.3570$\stdv{0.0013} \\
\bottomrule
\end{tabular}}
\label{tab:lambda}
\end{table}

Table~\ref{tab:lambda} sweeps the coverage weight on the primary cells, reporting three quantities per cell: the usage entropy of the converged selection, the mean pairwise cosine among the selected anchor keys, and forecasting error. Error is lowest at $\lambda_{\text{cov}}=0.1$ on both ETTh1 cells and rises again at $\lambda_{\text{cov}}=0.5$, so the useful setting there is a moderate weight rather than the largest one. On both ETTm1 cells the four weights lie within seed spread of one another, and the tuned setting is the one selected by validation MSE rather than the one that minimizes test error.

The two selection statistics separate. Usage entropy stays at its floor of $\ln K$ in thirteen of the sixteen cells, including every $\lambda_{\text{cov}}=0$ cell: the converged model concentrates on a small fixed set of anchors whether or not the coverage term is present, and the term does not broaden that set. Key cosine moves over the same sweep, on ETTh1-96 from $0.9971$ at $\lambda_{\text{cov}}=0$ to $0.3657$ at the setting that minimizes error. The two conditions also settle on different anchors: the converged selection sets are near-disjoint across conditions, with Jaccard overlap near zero on most seeds, so coverage relocates the fixed selection to a different region of the pool rather than broadening it. The quantity the term changes is the arrangement of the selected keys, not how many anchors are selected or how evenly.

The effect is confined. On ETTh1-336 cosine stays above $0.92$ at every weight while error still improves by $4.2\%$, so the geometric reshaping seen at horizon 96 is neither necessary for the error reduction nor present at every horizon. Appendix~\ref{app:sensitivity:pca} reports the corresponding PCA view on ETTh1-96, and Appendix~\ref{app:sensitivity:concentration} reports the separate training-time picture, in which usage does redistribute across batches before converging. Appendix~\ref{app:significance:balancing} substitutes entropy regularization and load balancing for the coverage term on the same cells: neither spreading the usage histogram nor holding it fixed lowers error.

\begin{figure}[h!]
\centering
\includegraphics[width=0.85\columnwidth]{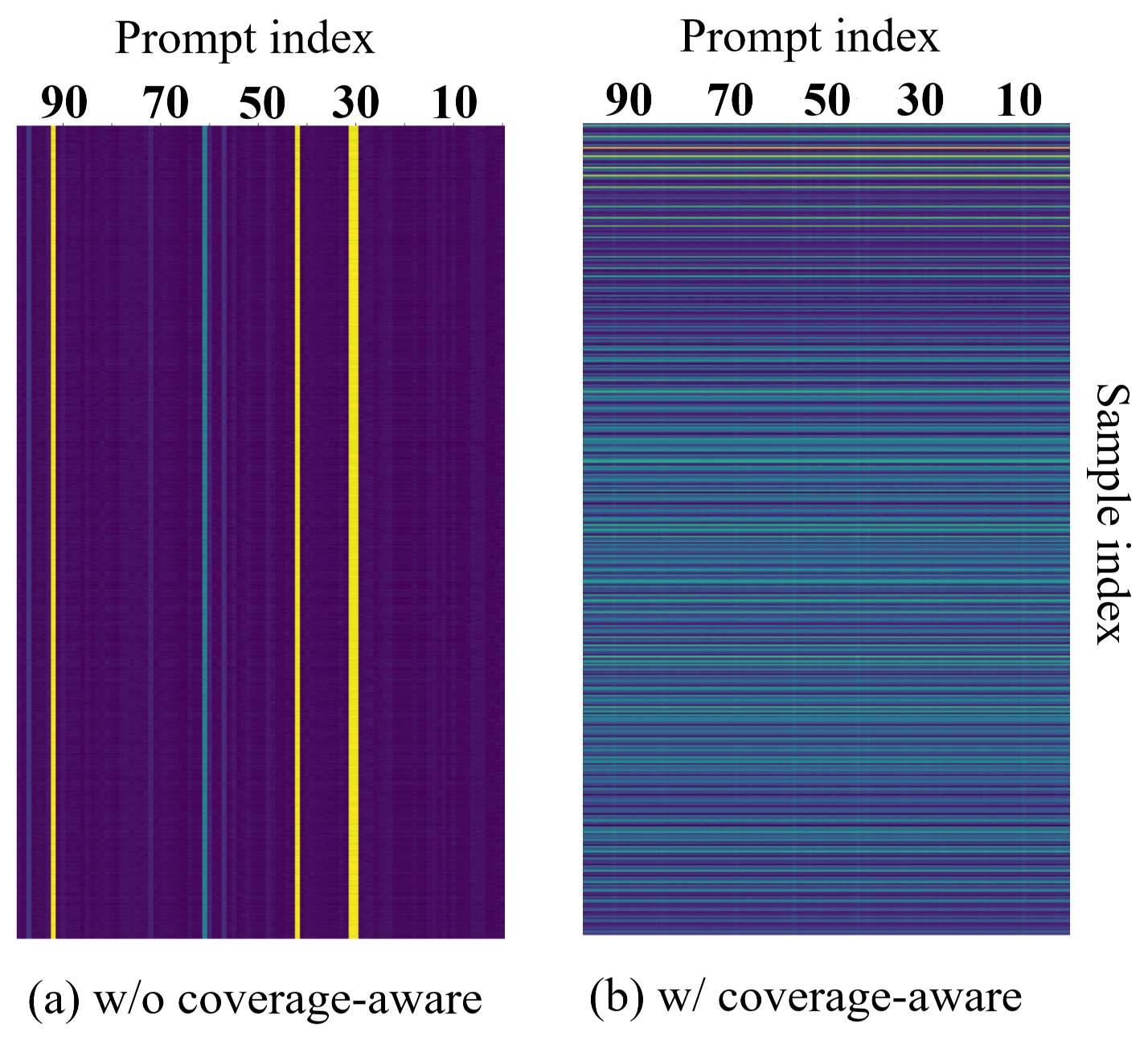}
\caption{Selection similarity score distributions under different prompt selection mechanisms.}
\label{fig:selection_score}
\end{figure}

Figure~\ref{fig:selection_score} visualizes the distribution of selection similarity scores for the top-100 anchors prior to top-$K$ selection. Without coverage-aware selection (panel a), a small set of anchor indices consistently maintain high scores across samples, forming clear horizontal bands. With coverage-aware selection (panel b), the dominant horizontal patterns weaken and scores disperse across broader anchor indices, while certain anchors retain high scores over specific sample segments, indicating that relevance-driven selection remains intact while high scores spread over a wider set of anchors. The banding in panel (a) indicates collapse onto a narrow set of preferred embedding directions, the bias regime our coverage objective targets; the dispersion in panel (b) reflects structural reorganization rather than mere score rescaling, since uniform attenuation would shift band magnitudes without altering their shape.

\begin{table}[h!]
\centering
\small
\caption{Ablation on prompt alignment (P+A) and decomposition (D). Single run per cell.}
\label{tab:ablation_prompt_decomp}
\resizebox{\columnwidth}{!}{%
\begin{tabular}{l||c|c|c|c}
\toprule
 & \textbf{ETTh1-96} & \textbf{ETTh1-192} & \textbf{ETTm1-96} & \textbf{ETTm1-192} \\
\midrule
w/o P+A, w/o D & 0.3760 & 0.4193 & 0.2990 & 0.3424 \\
w/ P+A, w/o D  & 0.3742 & 0.4094 & 0.2983 & 0.3324 \\
w/ P+A, w/ D   & \textbf{0.3671} & \textbf{0.3973} & \textbf{0.2917} & \textbf{0.3290} \\
\bottomrule
\end{tabular}%
}
\end{table}

\begin{table*}[h!]
\centering
\small
\caption{Anchor replacement ablation. We replace the retrieved semantic anchors with similarity-shuffled or randomly sampled anchors, holding all other components fixed.}
\label{tab:anchor_ablation}
\resizebox{\textwidth}{!}{%
\begin{tabular}{l||c|c|c|c|c|c|c|c}
\toprule
 & \textbf{ETTh1-96} & \textbf{ETTh1-192} & \textbf{ETTh2-96} & \textbf{ETTh2-192} & \textbf{ETTm1-96} & \textbf{ETTm1-192} & \textbf{ETTm2-96} & \textbf{ETTm2-192} \\
\midrule
Full (Ours)       & \textbf{0.3671} & \textbf{0.3973} & \textbf{0.2776} & \textbf{0.3422} & \textbf{0.2917} & \textbf{0.3290} & \textbf{0.1642} & \textbf{0.2235} \\
Shuffled anchor   & 0.3739 & 0.3962 & 0.2801 & 0.3489 & 0.2944 & 0.3351 & 0.1659 & 0.2274 \\
Random anchor     & 0.3812 & 0.4051 & 0.2798 & 0.3561 & 0.3024 & 0.3318 & 0.1698 & 0.2289 \\
\bottomrule
\end{tabular}%
}
\end{table*}

Table~\ref{tab:ablation_prompt_decomp} reports an ablation on prompt alignment (P+A) and decomposition (D). Coverage-aware prompt alignment alone yields modest improvements over the reference (w/o P+A, w/o D); combining it with decomposition produces larger gains, indicating that decomposition-based temporal regularization is important for extracting structure from semantically aligned prompts.

\begin{table}[h!]
\centering
\scriptsize
\setlength{\tabcolsep}{4pt}
\caption{Coverage crossed with decomposition on ETTh1-96.}
\label{tab:crossed}
\resizebox{\columnwidth}{!}{%
\begin{tabular}{l|c||c|c|c|c}
\toprule
Input & Head & $\lambda_{\text{cov}}=0$ & $\lambda_{\text{cov}}=0.1$ & Improved & $p$ \\
\midrule
Decomposed & 3-way & $0.3726$\stdv{0.0050} & $\mathbf{0.3702}$\stdv{0.0026} & 7/8 & $0.039$ \\
Raw & 3-way & $\mathbf{0.3663}$\stdv{0.0029} & $0.3717$\stdv{0.0058} & 1/8 & $0.023$ \\
Raw & 1-way & $0.3753$\stdv{0.0026} & $0.3753$\stdv{0.0027} & 3/8 & $1.000$ \\
\bottomrule
\end{tabular}}
\end{table}

Table~\ref{tab:crossed} crosses the coverage term with the decomposition on the primary cell, holding the prompt path active throughout. The coverage term lowers error only in the decomposed row, improving seven of eight seeds. With raw input the effect does not carry: it is absent when the output head is collapsed, and it reverses when the head is kept. The term is therefore coupled to the decomposed representation rather than acting as a general regularizer that survives a change of input.

The two raw rows also show that removing the decomposition is not a single intervention. Collapsing the head along with the input triples the output layer and gives the weakest configuration of the three, while keeping the head matches the full model's parameter count and gives the lowest error in the comparison at $\lambda_{\text{cov}}=0$. On this cell the decomposed input is not what drives accuracy; it is the representation through which the coverage term acts, and removing it removes the term's gain.
 We further test whether the retrieved anchors carry semantic information by replacing the top-$K$ anchors with similarity-shuffled or randomly sampled ones from the pool (Table~\ref{tab:anchor_ablation}). Performance degrades in 15 of the 16 setting-manipulation pairs, and random replacement gives the larger average degradation (+2.6\% against +1.2\% for shuffling), indicating that the retrieved anchors carry semantic information that contributes beyond embedding-space proximity alone.

\begin{figure}[h!]
\centering
\includegraphics[width=\columnwidth]{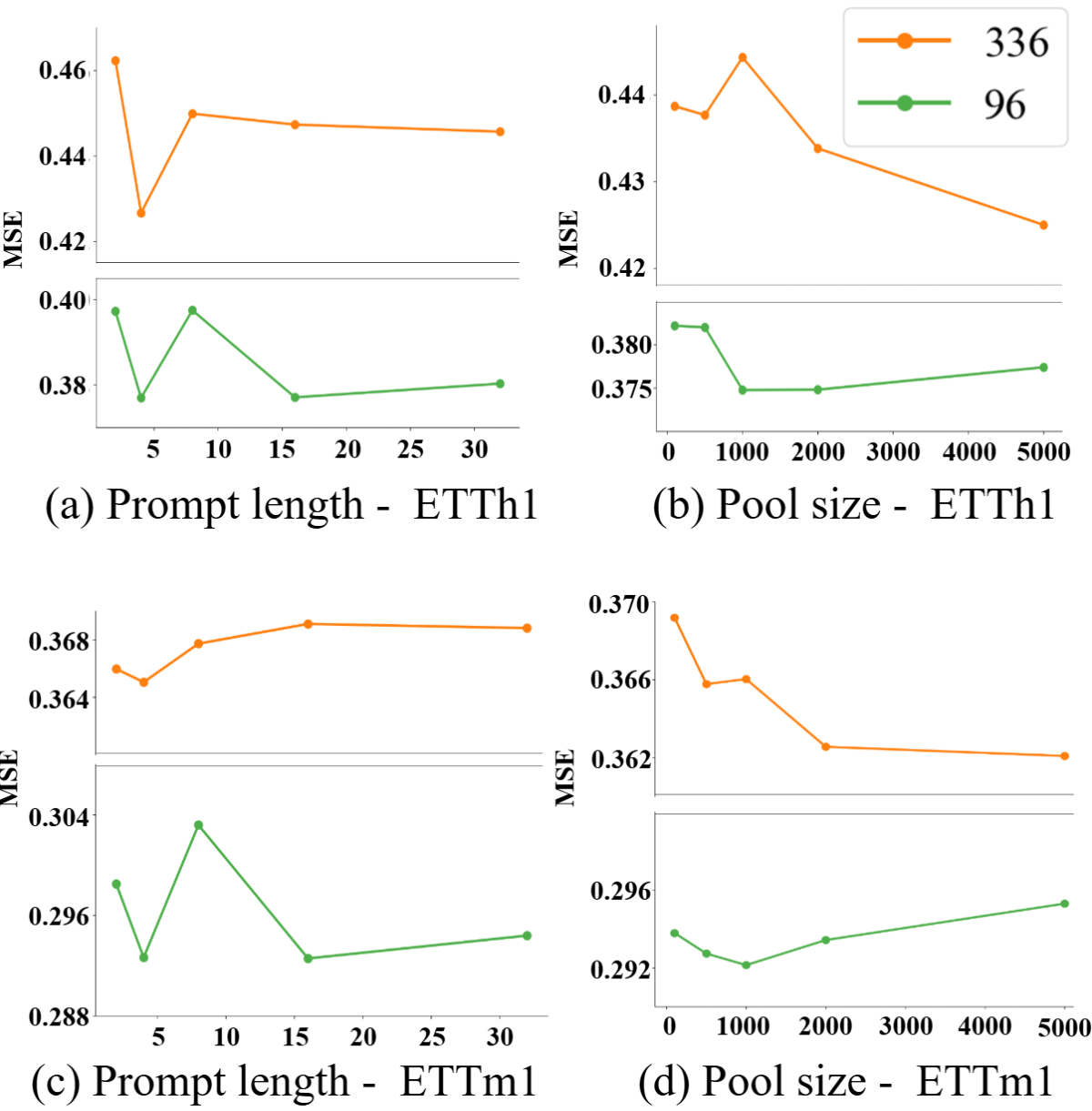}
\caption{Ablation on prompt length and pool size. Single run per cell.}
\label{fig:ablation_all}
\end{figure}

Figure~\ref{fig:ablation_all} reports sensitivity to prompt length and anchor pool size. Prompt length is non-monotonic: error is lowest at $K=4$ on all four curves, the 96-step horizons spike at $K=8$ and recover at 16, and the 336-step horizons rise mildly after 4 and flatten. Larger pools lower error at horizon 336 on both datasets, whereas at horizon 96 error is lowest at 1000 anchors and rises again beyond 2000.

\begin{figure*}[h!]
\centering
    \includegraphics[width=\textwidth]{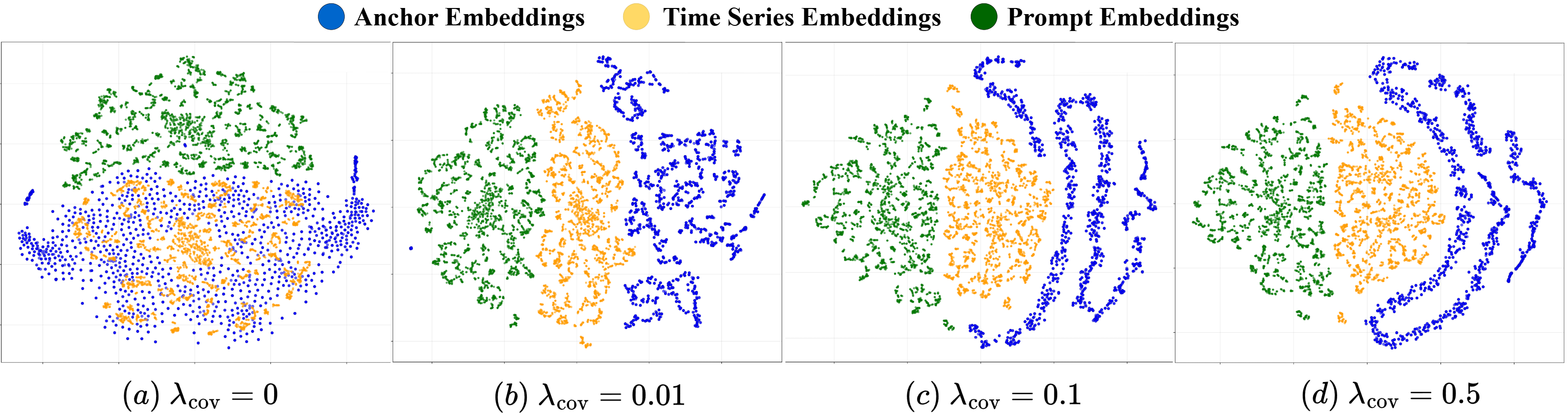}
    \caption{t-SNE visualization of semantic anchors (blue), time series embeddings (orange), and prompt embeddings (green) under different coverage loss coefficients $\lambda_{\text{cov}}$. As $\lambda_{\text{cov}}$ increases from 0 to 0.5, anchor embeddings become more dispersed, leading to a more balanced alignment of time series and prompt embeddings across the anchor space.}
    \label{fig:tsne_cov_ablation}
\end{figure*}

To qualitatively assess how coverage-aware selection restructures the representation space, Figure~\ref{fig:tsne_cov_ablation} visualizes semantic anchors (blue), raw time-series embeddings (orange), and prefix-prompted embeddings (green) under different values of $\lambda_{\text{cov}}$. At $\lambda_{\text{cov}} = 0$, anchor embeddings form dense clusters and the time-series and prefix embeddings are entangled. As $\lambda_{\text{cov}}$ increases, the anchors disperse more widely in the projection and prefix embeddings form clearer clusters; at $\lambda_{\text{cov}} = 0.5$, anchors exhibit well-separated boundaries and prefix embeddings bridge between time-series and anchors. Additional embedding alignment maps are reported in Appendix~\ref{app:qualitative}.

%% file: 05_discussion.tex
\section{Discussion}
\label{sec:discussion}

\subsection{Connections to Retrieval and Coverage Mechanisms}
\label{sec:discussion:connections}

The coverage criterion in CASP-LLM shares conceptual ground with diversity-aware in-context example selection~\cite{gupta2023coverage, liu2024se2, ye2023complementary}, which addresses the redundancy of similarity-only retrieval through submodular optimization~\cite{gupta2023coverage}, determinantal point processes~\cite{kulesza2012determinantal, ye2023complementary}, and sequential refinement~\cite{liu2024se2}. CASP-LLM differs in two respects: the coverage signal is accumulated across batches through an EMA-based usage statistic rather than computed within a single retrieval call, preserving cross-batch memory of selection patterns and integrating with the training objective; and the retrieval target is a continuous semantic anchor pool derived from pretrained LLM embeddings rather than a discrete labeled exemplar pool. We expand on the connection to RAG-style passage re-ranking and to NMT coverage in Appendix~\ref{app:coverage_connections}.

To test whether within-sample selection diversification improves retrieval, we replace the relevance-only top-$K$ rule with maximal marginal relevance (MMR)~\cite{carbonell1998use}, which trades off query relevance against pairwise anchor redundancy under a coefficient $\alpha$, and evaluate $\alpha \in \{0.5, 0.7\}$ over three seeds. On ETTh1-96, aggressive diversification ($\alpha = 0.5$) degrades MSE by $2.2\%$ relative to the relevance-only baseline, while moderate diversification ($\alpha = 0.7$) stays within $0.1\%$ of it; on ETTm1-96 both settings remain within $0.1\%$ of the baseline. A five-seed extension on ETTh2-96 and ETTm2-96 with $\alpha \in \{0.3, 0.5, 0.7\}$ gives the same picture: every setting stays within one sample standard deviation of the relevance-only baseline, and the per-seed difference changes sign across seeds. No MMR configuration improves over relevance-only selection on any of the four cells. We read this as evidence that redundancy among the anchors retrieved for a single forecast, the axis underlying MMR and related re-rankers, is not the operative mechanism for this failure mode: penalizing it either hurts accuracy or leaves it unchanged. This motivates regularizing at the cross-batch usage level instead, where the coverage statistic accumulates selection history across training rather than diversifying a single retrieval call. A broader comparison across datasets and horizons, and of hybrid mechanisms that combine within-sample and cross-batch signals, is left to future work.

\subsection{Portability Beyond Time Series Forecasting}
\label{sec:discussion:portability}

The coverage mechanism is not specific to forecasting. Wherever an LLM is conditioned on retrieved exemplars or prototype embeddings from a learned pool, similarity-only retrieval produces the redundancy issue documented in Section~\ref{sec:related:retrieval} and a usage-aware re-ranking objective is applicable. The most direct candidates are language settings where prior work has already established the value of diversity-aware retrieval: in-context example selection over labeled exemplars~\cite{ye2023complementary, gupta2023coverage}, passage retrieval in RAG~\cite{lewis2020retrieval}, and prompt-pool methods in continual learning~\cite{wang2022l2p}. The principle is also portable to other non-textual sequence modalities such as audio, motion, or sensor data, wherever the same retrieval-conditioning pattern is used. Empirical validation across modalities is left to future work.

%% file: 07_conclusion.tex
\section{Conclusion}
\label{sec:conclusion}

We propose CASP-LLM, a coverage-aware semantic prompting framework that addresses prompt selection bias in LLM-based time series forecasting by combining usage-tracking and a saturating gate into a regularizer that adds no learnable parameters, linking LLM-for-TS forecasting to the broader prompt-retrieval literature. Across benchmarks CASP-LLM is competitive with strong baselines, and the paired comparison against the same model without the coverage term isolates where the mechanism acts: it controls anchor usage at the cross-batch level rather than per-retrieval. The effect is absent on Electricity and in the few-shot long-horizon regime.

%% file: 06_limitations.tex
\section{Limitations}
\label{sec:limitations}

Our empirical validation is restricted to long- and short-horizon time-series forecasting; transfer of the mechanism to in-context learning, retrieval-augmented generation, or other non-textual modalities is left to future work. The coverage mechanism is integrated tightly with the semantic anchor pool from pretrained LLM word embeddings, and disentangling its contribution from that of the anchor design (e.g., against learned soft prompts without semantic grounding) requires further controlled comparison. We evaluate a single LLM backbone (GPT-2). Equation~(\ref{eq:cov_grad}) fixes the shape of the coverage pressure but not its endpoint, and the interaction between the selection budget $K$, the anchor pool size $V'$, and the batch size is characterized empirically rather than theoretically; a pool small relative to the batch, where the per-anchor statistic is updated from few distinct queries, is the regime we would expect to be most sensitive. The few-shot setting marks a related boundary (Appendix~\ref{app:fewshot}): at 10\% training data and horizon 720, CASP-LLM trails $\text{S}^2$IP-LLM and OFA on every ETT dataset, and trails DLinear on ETTm1 (0.487 against 0.411) and ETTm2 (0.366 against 0.316). This configuration supplies the fewest training windows we evaluate, so the usage statistic is least converged in this setting. Whether a warm-up or a longer EMA horizon closes this gap is open, as are the two stratification axes specified in Appendix~\ref{app:significance:axes}.

%% file: 08_app_setup.tex
\section{Detailed Experimental Setup}
\label{app:setup}

\subsection{Dataset Descriptions}
\label{app:setup:datasets}

For long-term forecasting, we evaluate CASP-LLM on six widely-used multivariate time series datasets. Five are standard benchmarks: \textit{Weather} (21 meteorological indicators, 10-minute intervals), \textit{ETTh1} and \textit{ETTh2} (electricity transformer temperature, hourly), and \textit{ETTm1} and \textit{ETTm2} (electricity transformer temperature, 15-minute). We additionally include the higher-dimensional \textit{Electricity} dataset, comprising 321 electricity consumption channels with hourly resolution, to evaluate scalability and robustness under high-dimensional and strongly seasonal conditions. Each dataset exhibits distinct temporal characteristics, including trend, seasonality, and noise patterns. For short-term forecasting, we use the \textit{M4 Competition} benchmark~\cite{makridakis2020m4} containing 100{,}000 time series across six categories (Yearly, Quarterly, Monthly, Weekly, Daily, Hourly) that encompass data sparsity, irregular patterns, and varying seasonal components. Dataset statistics are summarized in Table~\ref{tab:dataset_statistics}.

\begin{table}[h!]
\scriptsize
\centering
\setlength{\tabcolsep}{3pt}
\resizebox{\columnwidth}{!}{%
\begin{tabular}{@{}l||l|c|c|c|c@{}}
\toprule
 & Dataset & Channels & Forecast Horizon & Length / \# Series & Freq. \\
\midrule
\multirow{6}{*}{\rotatebox{90}{Long-term}}
& Weather     & 21  & \{96, 192, 336, 720\} & 52696 & 10 min \\
& ETTh1       & 7   & \{96, 192, 336, 720\} & 17420 & 1 hour \\
& ETTh2       & 7   & \{96, 192, 336, 720\} & 17420 & 1 hour \\
& ETTm1       & 7   & \{96, 192, 336, 720\} & 69680 & 15 min \\
& ETTm2       & 7   & \{96, 192, 336, 720\} & 69680 & 15 min \\
& Electricity & 321 & \{96, 192, 336, 720\} & 26304 & 1 hour \\
\midrule
\multirow{6}{*}{\rotatebox{90}{Short-term}}
& Yearly (M4)    & 1 & 6  & 23000 & yearly \\
& Quarterly (M4) & 1 & 8  & 24000 & quarterly \\
& Monthly (M4)   & 1 & 18 & 48000 & monthly \\
& Weekly (M4)    & 1 & 13 & 359   & weekly \\
& Daily (M4)     & 1 & 14 & 4227  & daily \\
& Hourly (M4)    & 1 & 48 & 414   & hourly \\
\bottomrule
\end{tabular}}
\caption{Statistics of datasets in long-term and short-term forecasting tasks.}
\label{tab:dataset_statistics}
\end{table}

\subsection{Experimental Configuration}
\label{app:setup:config}

All experiments are conducted using the unified evaluation framework of \citet{wu2023timesnet}, implemented via the Time-Series-Library repository (\url{https://github.com/thuml/Time-Series-Library}). Training uses NVIDIA L40S GPUs on a server with an AMD EPYC 9654 processor (96 cores) and 1\,TB of memory; up to 4 GPUs are used for the Electricity dataset at the 720-step horizon and up to 2 GPUs for all other configurations. All datasets used in this work are publicly available forecasting benchmarks released for research use, and our implementation builds on the Time-Series-Library (MIT License); our use of these datasets and code is consistent with their intended research use.

CASP-LLM employs a GPT-2 decoder with 6 transformer layers, 768 hidden dimensions, and 12 attention heads. The semantic embedding dimension is $d = 768$, matching the pretrained word-token embedding width, with patch length $P_{\text{patch}} = 16$ and stride 8. We use the AdamW optimizer with weight decay $1\!\times\!10^{-5}$, train for up to 100 epochs with early stopping (patience 3), and repeat each experiment 3 times with different random seeds. The learning rate is $1\!\times\!10^{-4}$ for all configurations except Weather (all horizons) and ETTh1 at horizon 720, which use $1\!\times\!10^{-3}$ (Table~\ref{tab:hparams}). Batch sizes by dataset are 256 (Weather), 512 (Electricity), 128 (ETTh2, ETTm1, ETTm2), 64 (ETTh1), and 32 (M4 splits). The semantic anchor pool size is fixed at $V' = 1000$. In total, CASP-LLM has 134.0M parameters, of which 52.9M ($39.5\%$) are trainable. The trainable count is dominated by the semantic anchor projection $f$ (50.3M), which maps the $V = 50{,}257$ pretrained GPT-2 word-token embeddings to the $V' = 1000$ anchors via the learnable linear projection introduced in Section~\ref{sec:method}, following the anchor design of $\text{S}^2$IP-LLM~\cite{pan2024s2ip}. The remaining trainable parameters comprise the linear forecasting head (1.8M), the patch-embedding projection ($\sim$38K), and, within the GPT-2 backbone, only the positional embeddings and layer-normalization parameters (0.81M). The remaining 81.1M are frozen: the pretrained word-token embedding table (38.6M), from which the anchors are derived, together with the multi-head attention (14.2M) and feed-forward (28.3M) blocks of the six backbone layers. This trainable footprint is comparable to that of the closest similarity-based baseline $\text{S}^2$IP-LLM, which adopts the same word-embedding-derived anchor projection. The coverage term adds an EMA update over the $V'$-dimensional usage vector and an elementwise gate on the similarity matrix that top-$K$ selection already computes; it adds no parameters and leaves the forward pass and inference unchanged, so per-run training time is comparable to the $\lambda_{\text{cov}}=0$ configuration at matched seeds.

For long-term benchmarks, the input sequence length is 512 and prediction horizons are $\{96, 192, 336, 720\}$. Hyperparameter selection follows a two-stage protocol. First, we adopt the trend and seasonal lengths of $\text{S}^2$IP-LLM~\cite{pan2024s2ip} as initial values. Second, we tune per dataset and horizon by validation MSE over the loss weights $\lambda_{\text{sim}}$ and $\lambda_{\text{cov}}$, the prompt length $K$, and the decomposition lengths. Table~\ref{tab:hparams} lists the selected configuration for every dataset--horizon pair together with the batch size and learning rate; selected values span $\lambda_{\text{sim}} \in \{0.01, 0.05, 0.1\}$, $\lambda_{\text{cov}} \in \{0.01, 0.05, 0.1, 0.15\}$, $K \in \{2, 4, 8, 16\}$, \textit{trend length} $\in \{24, 48, 96, 192\}$, and \textit{seasonal length} $\in \{12, 24, 48, 96, 192\}$.

The coverage ablation in Table~\ref{tab:lambda} sweeps $\lambda_{\text{cov}} \in \{0, 0.01, 0.1, 0.5\}$ on the primary cells. The sweep extends beyond the tuning grid at both ends to show behavior under no regularization and under over-regularization, so it is a sensitivity analysis rather than a second search.
For the EMA decay factor, we use $\beta = 0.99$ as the default.

For the short-term M4 datasets, prediction horizons of 6, 8, 18, 13, 14, and 48 are used for Yearly, Quarterly, Monthly, Weekly, Daily, and Hourly frequencies respectively, with lookback length set to twice the corresponding horizon. Most hyperparameters are held constant across frequencies: $\lambda_{\text{sim}} = 0.05$ and $\lambda_{\text{cov}} = 0.05$.
The trend length is 4 for all frequencies; the seasonal length is 8 for Monthly and 4 for the remaining frequencies; prompt length is selected from $\{2, 4\}$ per frequency.

\paragraph{Decomposition details}
We apply Reversible Instance Normalization~\cite{kim2021revin} before decomposition. Given a univariate time series $X_{i,t}$ at time $t$ for variable $i$, the normalized value is
\begin{equation*}
X'_{i,t} = \gamma_T \cdot \frac{X_{i,t} - \mathbb{E}_{t}[X_{i,t}]}{\sqrt{\mathrm{Var}_t[X_{i,t}] + \epsilon_T}} + \beta_T,
\end{equation*}
where $\gamma_T$ and $\beta_T$ are learnable scale and shift parameters and $\epsilon_T$ is a small constant. We then apply the classical additive seasonal-trend decomposition adopted by $\text{S}^2$IP-LLM~\cite{pan2024s2ip} to obtain $X'_{i,t} = T_{i,t} + S_{i,t} + R_{i,t}$. The trend $T_{i,t}$ is a centered moving average of window $L_{\text{tr}}$; the seasonal component $S_{i,t}$ is the average of the detrended series $X'_{i,t} - T_{i,t}$ over time indices sharing the same position within a season of length $L_{\text{se}}$; and $R_{i,t}$ is the remainder. Positions left undefined by the centered window at the series boundary are filled by nearest-value extension. Both $L_{\text{tr}}$ and $L_{\text{se}}$ are tuned per dataset and horizon and are listed in Table~\ref{tab:hparams}; the ranges searched extend those used by $\text{S}^2$IP-LLM.

After decomposition, each component is segmented into overlapping patches as in PatchTST~\cite{nie2023patchtst}:
\begin{equation*}
P^{\text{trend}}_{i,t-\tau:t-1} = \left[X^{\text{trend}}_{i,t-\tau}, \dots, X^{\text{trend}}_{i,t-1}\right] \in \mathbb{R}^{N_p \times L_p},
\end{equation*}
where $L_p$ is the patch length and $N_p = \lfloor (\tau - L_p)/s \rfloor + 1$ is the patch count with stride $s$. The three components are concatenated along the channel dimension to form a composite patch $P^{\text{meta}}_{i,t-\tau:t-1} \in \mathbb{R}^{N_p \times 3 L_p}$, which is projected into the LLM embedding space via $\mathrm{Embed}(\cdot)$ to yield $Z_{i,t-\tau:t-1} \in \mathbb{R}^{N_p \times d}$.

\subsection{Baselines}
\label{app:setup:baselines}

We compare against the following established baselines.
\begin{itemize}
\item \textbf{PatchTST}~\cite{nie2023patchtst}: a Transformer that segments time series into patch sequences and applies self-attention to capture temporal dependencies.
\item \textbf{iTransformer}~\cite{liu2023itransformer}: inverts the attention axes to model long-range dependencies across variates.
\item \textbf{FEDformer}~\cite{zhou2022fedformer}: a frequency-enhanced decomposition Transformer capturing global and local temporal patterns.
\item \textbf{DLinear}~\cite{zeng2023dlinear}: a streamlined linear baseline with strong empirical performance.
\item \textbf{OFA}~\cite{zhou2023one}: a unified framework that feeds patched time-series tokens directly into a frozen GPT-2, fine-tuning only the positional embedding and layer-normalization parameters, without natural-language prompts.
\item \textbf{Time-LLM}~\cite{jin2024time}: reprograms patch embeddings into text prototypes and concatenates declarative natural-language prompts.
\item \textbf{CALF}~\cite{liu2025calf}: cross-modal fine-tuning with distribution alignment between textual and temporal embeddings.
\item \textbf{$\text{S}^2$IP-LLM}~\cite{pan2024s2ip}: semantic anchor pool retrieval from pretrained GPT-2 word embeddings; the closest similarity-based prior method to CASP-LLM.
\end{itemize}

FEDformer and Autoformer~\cite{wu2021autoformer} are additionally included as non-LLM baselines in the short-term M4 comparison (Table~\ref{tab:short_full}); the long-term comparison in Table~\ref{tab:longterm} focuses on the LLM-based methods and the strongest recent non-LLM models.

All baseline numbers are obtained under the unified evaluation protocol described above, using official implementations where available. Reported values may therefore differ from those in the original papers due to differences in the evaluation setup, and all methods, including CASP-LLM, are evaluated under identical conditions to ensure that comparisons are internally consistent.

\paragraph{Scope of the prompt-pool comparison}
TEMPO~\cite{cao2023tempo} belongs to the same retrieval-conditioned family as $\text{S}^2$IP-LLM and applies the same top-$K$ cosine rule, so it is a natural target for a coverage term rather than a design outside its reach. The comparisons in this paper are paired: the $\lambda_{\text{cov}}=0$ and coverage conditions share code, data, seeds, and tuned hyperparameters, which is what allows a per-cell difference to be read as an effect of the coverage term (Appendix~\ref{app:significance}). Meeting that standard on TEMPO involves an injection point for the usage statistic matched to where selection occurs in its trend, seasonal, and residual prompt construction, a retuning of the coverage weight and prompt length under TEMPO's own configuration, and paired seed runs at the same budget as Table~\ref{tab:paired_cov0}. A port without that retuning would evaluate the coverage term against a configuration never tuned for it, so we report the anchor-based instantiation and leave the prompt-bank instantiation open.

\subsection{Per-Dataset Hyperparameters}
\label{app:setup:hparams}

Hyperparameters selected by validation MSE are listed per dataset and horizon in Table~\ref{tab:hparams}, together with the batch size and learning rate used for each configuration. Every comparison that isolates the coverage term (Tables~\ref{tab:lambda}, \ref{tab:paired_cov0}, and \ref{tab:balancing}) varies only the coverage weight or objective within the configuration listed here, so this tuning is held fixed across the compared conditions.

\begin{table}[h!]
\centering
\scriptsize
\setlength{\tabcolsep}{3pt}
\resizebox{\columnwidth}{!}{%
\begin{tabular}{l|c||c|c|c|c|c|c|c}
\toprule
Dataset & $H$ & $\lambda_{\text{sim}}$ & $\lambda_{\text{cov}}$ & $K$ & Trend & Seas. & Batch & LR \\
\midrule
\multirow{4}{*}{ETTh1}   & 96  & 0.05 & 0.1  & 8  & 96  & 96  & 64  & $10^{-4}$ \\
                         & 192 & 0.05 & 0.1  & 8  & 96  & 12  & 64  & $10^{-4}$ \\
                         & 336 & 0.1  & 0.1  & 8  & 96  & 12  & 64  & $10^{-4}$ \\
                         & 720 & 0.01 & 0.1  & 4  & 48  & 48  & 64  & $10^{-3}$ \\
\midrule
\multirow{4}{*}{ETTh2}   & 96  & 0.1  & 0.05 & 16 & 96  & 12  & 128 & $10^{-4}$ \\
                         & 192 & 0.1  & 0.05 & 4  & 96  & 12  & 128 & $10^{-4}$ \\
                         & 336 & 0.1  & 0.05 & 8  & 96  & 12  & 128 & $10^{-4}$ \\
                         & 720 & 0.1  & 0.05 & 8  & 96  & 192 & 128 & $10^{-4}$ \\
\midrule
\multirow{4}{*}{ETTm1}   & 96  & 0.05 & 0.15 & 4  & 96  & 96  & 128 & $10^{-4}$ \\
                         & 192 & 0.05 & 0.1  & 4  & 192 & 96  & 128 & $10^{-4}$ \\
                         & 336 & 0.05 & 0.01 & 4  & 192 & 96  & 128 & $10^{-4}$ \\
                         & 720 & 0.05 & 0.01 & 8  & 192 & 192 & 128 & $10^{-4}$ \\
\midrule
\multirow{4}{*}{ETTm2}   & 96  & 0.05 & 0.05 & 4  & 96  & 24  & 128 & $10^{-4}$ \\
                         & 192 & 0.05 & 0.1  & 8  & 96  & 48  & 128 & $10^{-4}$ \\
                         & 336 & 0.05 & 0.1  & 8  & 96  & 48  & 128 & $10^{-4}$ \\
                         & 720 & 0.05 & 0.1  & 8  & 96  & 48  & 128 & $10^{-4}$ \\
\midrule
\multirow{4}{*}{Weather} & 96  & 0.1  & 0.15 & 2  & 96  & 48  & 256 & $10^{-3}$ \\
                         & 192 & 0.1  & 0.1  & 2  & 96  & 48  & 256 & $10^{-3}$ \\
                         & 336 & 0.1  & 0.1  & 4  & 96  & 48  & 256 & $10^{-3}$ \\
                         & 720 & 0.1  & 0.1  & 4  & 96  & 48  & 256 & $10^{-3}$ \\
\midrule
\multirow{4}{*}{Electricity} & 96 & 0.1 & 0.1 & 4 & 24 & 48 & 512 & $10^{-4}$ \\
                         & 192 & 0.1  & 0.1  & 4  & 24  & 48  & 512 & $10^{-4}$ \\
                         & 336 & 0.1  & 0.1  & 4  & 24  & 24  & 512 & $10^{-4}$ \\
                         & 720 & 0.1  & 0.1  & 4  & 24  & 24  & 512 & $10^{-4}$ \\
\bottomrule
\end{tabular}}
\caption{Selected hyperparameters per dataset and horizon.}
\label{tab:hparams}
\end{table}

\paragraph{Code release}
Code and configuration files for all experiments are available at \url{https://github.com/dadaeun09/CASP-LLM}.

\subsection{Evaluation Metrics}
\label{app:setup:metrics}

For long-term forecasting we report Mean Squared Error (MSE) and Mean Absolute Error (MAE). For short-term M4 evaluation we report symmetric Mean Absolute Percentage Error (sMAPE), Mean Absolute Scaled Error (MASE), and Overall Weighted Average (OWA, the official M4 composite metric):
\begin{equation*}
\text{MSE} = \frac{1}{H} \sum_{h=1}^{H} (Y_h - \hat{Y}_h)^2, \quad
\text{MAE} = \frac{1}{H} \sum_{h=1}^{H} |Y_h - \hat{Y}_h|,
\end{equation*}
\begin{equation*}
\text{sMAPE} = \frac{200}{H} \sum_{h=1}^{H} \frac{|Y_h - \hat{Y}_h|}{|Y_h| + |\hat{Y}_h|}
\end{equation*}
\begin{equation*}
\text{MASE} = \frac{1}{H} \sum_{h=1}^{H} \frac{|Y_h - \hat{Y}_h|}{\frac{1}{n-s} \sum_{j=s+1}^{n} |Y_j - Y_{j-s}|},
\end{equation*}
\begin{equation*}
\text{OWA} = \frac{1}{2}\left(\frac{\text{sMAPE}}{\text{sMAPE}_{\text{Na\"{i}ve2}}} + \frac{\text{MASE}}{\text{MASE}_{\text{Na\"{i}ve2}}}\right),
\end{equation*}
where $H$ is the forecast horizon, $n$ is the length of the observed series, $s$ is the seasonal period, and $Y_h$, $\hat{Y}_h$ denote actual and predicted values at step $h$.

%% file: 09_app_mechanism.tex
\section{Coverage Mechanism Design Rationale}
\label{app:mechanism}

This appendix elaborates on the design choices in Section~\ref{sec:method:coverage}. Each subsection isolates one component, states the design intent, and cites the related literature.

\subsection{Anchor Usage Statistic}
\label{app:mechanism:ema}

The coverage statistic $\tilde{c}_i$ in Equation~(\ref{eq:cov_update}) is maintained as an exponential moving average over the per-batch mean similarity. The saturating $\min$ gate requires $\tilde{c}_i$ and the gated similarity $s_{b,i}$ to lie on the same numerical scale throughout training: if $\tilde{c}_i$ drifts much above the similarity range, the gate degenerates to $s_{b,i}$ and the usage-balancing signal vanishes; if it drifts much below, the gate uniformly suppresses all anchors and the relevance signal is lost. An EMA satisfies this scale-matching property by construction. With decay $\beta \in (0,1)$ and $\bar{s}_i^{(t)} = (1/B) \sum_b \max(s_{b,i}^{(t)}, 0)$, the stationary mean of $\tilde{c}_i^{(t)}$ approximately matches $\mathbb{E}[\max(s_{b,i}, 0)]$, which is exactly the scale of the gated quantity in Equation~(\ref{eq:cov_loss}). The saturating gate therefore remains informative throughout training without ad-hoc rescaling.

EMA tracking of a bounded running statistic is a common pattern for this role in deep learning, including target tracking in mean-teacher consistency training~\cite{tarvainen2017meanteacher} and prompt-pool continual learning~\cite{wang2022l2p}; the underlying goal of balancing usage across a pool also appears in expert-routing load balancing for mixture-of-experts~\cite{fedus2022switch}. We adopt the construction here because it adapts smoothly to non-stationary anchor embeddings during training, has a single intuitive hyperparameter ($\beta$, with effective forgetting horizon $1/(1-\beta)$), and inherits the gradient stability properties of EMA-based targets in semi-supervised learning.

\subsection{Saturating min Gate}
\label{app:mechanism:min}

Given a bounded usage statistic $\tilde{c}_i$, several functional forms can implement a usage-aware regularizer. The $\min$ operator implements the canonical saturating-gate semantics: when an anchor is under-utilized ($\tilde{c}_i$ small), the gate is bounded by $\tilde{c}_i$ and the optimization signal through anchor $i$ is attenuated, encouraging redistribution. Once the anchor has accumulated sufficient usage, $\tilde{c}_i \geq s_{b,i}$, the gate degenerates to $s_{b,i}$, and the optimization signal flows freely. This is the same saturation-threshold behavior used in submodular coverage objectives~\cite{lin2011submodular} and in maximal marginal relevance~\cite{carbonell1998use}, with the difference that $\min$ operates at the per-item level rather than as a set-level constraint.

\subsection{Stop-Gradient on the Coverage Statistic}
\label{app:mechanism:stopgrad}

Equation~(\ref{eq:cov_loss}) applies a stop-gradient operator to $\tilde{c}_i$. Writing $u_{b,i} = \max(s_{b,i}, 0)$, the summand is $\min(u_{b,i}, \mathrm{sg}[\tilde{c}_i^{(t)}])$, and with $\tilde{c}_i^{(t)}$ treated as a constant the derivative is piecewise constant with three branches:
\begin{equation}
\label{eq:cov_grad}
\frac{\partial \mathcal{L}_{\text{cov}}}{\partial s_{b,i}} = \frac{1}{B} \, \mathds{1}\bigl[\, 0 < s_{b,i} < \tilde{c}_i^{(t)} \,\bigr].
\end{equation}
The three branches are: $s_{b,i} \leq 0$, where the clipping zeroes the derivative; $0 < s_{b,i} < \tilde{c}_i^{(t)}$, where the gate returns the similarity and a descent step lowers $s_{b,i}$; and $s_{b,i} \geq \tilde{c}_i^{(t)}$, where the gate saturates at a constant and the derivative vanishes. Two properties follow directly. First, the coverage term contributes at most $\lambda_{\text{cov}}/B$ to the gradient of any single similarity and at most $\lambda_{\text{cov}}$ summed over a batch, so it cannot dominate the forecasting gradient by scale alone. Second, the anchors that receive pressure at step $t$ are exactly $\mathcal{A}_b^{(t)} = \{ i : 0 < s_{b,i} < \tilde{c}_i^{(t)} \}$, which depends on $\tilde{c}_i^{(t)}$ through a threshold: an anchor whose accumulated affinity is high has a high threshold and is active for most queries, while an anchor with $\tilde{c}_i^{(t)}$ near zero is saturated for every query with positive similarity and receives nothing. Pressure is therefore differential in accumulated usage by construction rather than by tuning, and it is applied to the anchors that dominate retrieval while leaving rarely used anchors untouched.

Without the stop-gradient this structure disappears. The statistic $\tilde{c}_i^{(t)}$ depends on the current batch through Equation~(\ref{eq:cov_update}), with $\partial \tilde{c}_i^{(t)} / \partial s_{b,i} = (1-\beta)/B$ whenever $s_{b,i} > 0$. On the saturated branch the summand equals $\tilde{c}_i^{(t)}$, so differentiating through it contributes $(1-\beta)/B$ instead of zero, for every anchor and every sample with positive similarity. The loss would then admit a descent direction that lowers all similarities together, including those of anchors that are never selected, which reduces $\mathcal{L}_{\text{cov}}$ without changing the relative ordering that determines retrieval. The stop-gradient removes that direction, leaving relative reshaping of the similarity landscape as the way the term can be reduced. The same gradient-leakage hazard motivates the detached target in mean-teacher consistency training~\cite{tarvainen2017meanteacher}.

What Equation~(\ref{eq:cov_grad}) establishes is the shape and the scale of the pressure, not its endpoint. The threshold $\tilde{c}_i^{(t)}$ is itself a function of the parameters through past steps, so the active set moves during training and the objective is non-stationary; we give no convergence argument and no characterization of the stationary points. The behavior the mechanism does produce is measured rather than derived: at convergence the usage histogram sits near its floor in both conditions while the geometry of the selected anchor keys differs, which is the observable consequence reported in Section~\ref{sec:exp:ablation}.

\subsection{Training-Only Updates}
\label{app:mechanism:trainonly}

The EMA update to $\tilde{c}_i$ is gated on the training mode of the module and is disabled during validation and inference. This prevents two failure modes. First, during evaluation passes, applying the EMA update would drift $\tilde{c}_i$ toward statistics computed on held-out data, contaminating the gate used at subsequent training steps. Second, validation MSE is reported on a stationary state of the model and gate; if $\tilde{c}_i$ continues to evolve during evaluation, the reported validation curve becomes non-stationary in a way that complicates early stopping. By disabling updates outside training, the validation metric depends only on weights, not on the bookkeeping state.

\subsection{Summary}
\label{app:mechanism:summary}

Table~\ref{tab:design_summary} consolidates the design choices discussed above, the specific failure mode each one prevents, and its closest precedent in the literature.

\begin{table*}[h!]
\centering
\small
\begin{tabular}{l|l|l}
\toprule
Design choice & Failure mode prevented & Precedent \\
\midrule
EMA usage statistic       & Scale drift between usage and similarity & \cite{tarvainen2017meanteacher} \\
Saturating $\min$ gate    & Non-saturating regularization at high usage & \cite{lin2011submodular, carbonell1998use} \\
Stop-gradient on $\tilde{c}_i$ & Trivial similarity collapse              & \cite{tarvainen2017meanteacher} \\
Training-only update      & Evaluation-time drift of gate              & batch-norm convention \\
\bottomrule
\end{tabular}
\caption{Coverage mechanism design choices and their literature anchors.}
\label{tab:design_summary}
\end{table*}
\subsection{Extended Connections to Retrieval and Coverage Mechanisms}
\label{app:coverage_connections}

The coverage criterion connects to two lines of work that we summarize in Section~\ref{sec:discussion:connections} and expand here. First, retrieval-augmented generation systems~\cite{lewis2020retrieval} face an analogous redundancy problem: top-$K$ retrieval over a document corpus often returns near-duplicate passages, motivating diversity-aware re-ranking such as maximal marginal relevance~\cite{carbonell1998use}. Our mechanism can be viewed as an LLM-for-TS counterpart of this re-ranking, and applying the same usage-penalized scoring to passage retrieval in conventional RAG pipelines is a natural extension left to future work. Second, coverage in CASP-LLM relates to coverage mechanisms in neural sequence generation, originally developed to mitigate over- and under-translation in neural machine translation~\cite{tu2016modeling, mi2016coverage} and adapted for repetition control in summarization~\cite{see2017get}. Those mechanisms operate at the token-attention level during decoding and accumulate attention history within a single generation trajectory, whereas our coverage statistic operates at the retrieval level over a learned semantic anchor pool and accumulates usage across training batches; the underlying principle of penalizing repeated attention to the same elements is shared. CASP-LLM can therefore be read as a transposition of the NMT coverage idea from token-level attention to retrieval-pool-level usage.

%% file: 10_app_algorithm.tex
\section{Algorithm Pseudocode}
\label{app:algorithm}
Algorithm~\ref{alg:coverage} provides pseudocode for the coverage-aware selection module of Section~\ref{sec:method:coverage}. The procedure operates on the similarity matrix $\mathbf{s} \in \mathbb{R}^{B \times V'}$ between batch queries and anchor keys, and is invoked once per forward pass. We write $\mathrm{sg}[\cdot]$ for the stop-gradient operator.

\begin{algorithm}[h!]
\caption{Coverage-aware anchor selection.}
\label{alg:coverage}
\begin{algorithmic}[1]
\Require similarity matrix $\mathbf{s}\in\mathbb{R}^{B\times V'}$, EMA buffer $\tilde{\mathbf{c}}\in\mathbb{R}^{V'}$, decay $\beta$, top-$K$ size $K$, training flag $t$
\Ensure selected indices $\mathbf{I}$, coverage loss $\mathcal{L}_{\text{cov}}$, similarity loss $\mathcal{L}_{\text{sim}}$, updated buffer $\tilde{\mathbf{c}}$
\State $\mathbf{s}^{+} \gets \max(\mathbf{s},\,0)$ \Comment{align scale with $\tilde{\mathbf{c}}$}
\If{$t$ is true}
    \State $\bar{\mathbf{s}} \gets \tfrac{1}{B}\sum_{b=1}^{B}\mathbf{s}^{+}_{b,:}$ \Comment{per-anchor batch usage}
    \State $\tilde{\mathbf{c}} \gets \beta\,\tilde{\mathbf{c}} + (1-\beta)\,\mathrm{sg}[\bar{\mathbf{s}}]$ \Comment{EMA update}
\EndIf
\State $\mathbf{I} \gets \operatorname{TopK}(\mathbf{s},\,K)$ \Comment{greedy top-$K$ by raw similarity}
\State $\mathbf{G} \gets \min\bigl(\mathbf{s}^{+},\,\mathrm{sg}[\tilde{\mathbf{c}}]\bigr)$ \Comment{element-wise gating}
\State $\mathcal{L}_{\text{cov}} \gets \tfrac{1}{B}\sum_{b,v}\mathbf{G}_{b,v}$
\State $\mathbf{S}^{\text{sel}} \gets \operatorname{Gather}(\mathbf{s},\,\mathbf{I})$
\State $\mathcal{L}_{\text{sim}} \gets \tfrac{1}{B}\sum_{b,k}\bigl(1 - \mathbf{S}^{\text{sel}}_{b,k}\bigr)$
\State \Return $\mathbf{I},\,\mathcal{L}_{\text{cov}},\,\mathcal{L}_{\text{sim}},\,\tilde{\mathbf{c}}$
\end{algorithmic}
\end{algorithm}

The buffer $\tilde{\mathbf{c}}$ persists across forward passes and tracks per-anchor usage through the EMA in line~4. The stop-gradient on $\tilde{\mathbf{c}}$ in line~7 prevents the coverage loss from back-propagating through the EMA recursion, which would otherwise couple gradient signals across iterations (Appendix~\ref{app:mechanism:stopgrad}). For the MMR variant studied in Section~\ref{sec:discussion:connections}, line~6 is replaced with a greedy selection that augments each pick with a pairwise-redundancy penalty among already-selected anchors, controlled by the relevance-diversity coefficient $\alpha$; the coverage loss is then zero and the EMA update is skipped, so the only state carried across iterations is the anchor pool.

%% file: 11_app_longterm.tex
\section{Complete Long-term Forecasting Results}
\label{app:longterm}

Table~\ref{tab:full_results_avg} summarizes long-term forecasting results averaged across all four horizons $\{96, 192, 336, 720\}$ for each benchmark, complementing the per-horizon results reported in the main paper.

\begin{table*}[h!]
\centering
\caption{Average long-term forecasting results on benchmark datasets across all horizons. Lower MSE/MAE is better. Best in bold, second-best underlined.}
\label{tab:full_results_avg}
\scriptsize
\resizebox{\textwidth}{!}{%
\begin{tabular}{l|cc|cc|cc|cc|cc|cc|cc|cc}
\hline
\textbf{Dataset}
& \multicolumn{2}{c|}{\textbf{Ours}}
& \multicolumn{2}{c|}{\textbf{Time-LLM}}
& \multicolumn{2}{c|}{\textbf{S$^2$IP-LLM}}
& \multicolumn{2}{c|}{\textbf{CALF}}
& \multicolumn{2}{c|}{\textbf{OFA}}
& \multicolumn{2}{c|}{\textbf{iTransformer}}
& \multicolumn{2}{c|}{\textbf{PatchTST}}
& \multicolumn{2}{c}{\textbf{DLinear}} \\
\cline{2-17}
& MSE & MAE
& MSE & MAE
& MSE & MAE
& MSE & MAE
& MSE & MAE
& MSE & MAE
& MSE & MAE
& MSE & MAE \\
\hline
Weather     & \textbf{0.2261} & \textbf{0.2654} & 0.2346 & 0.2724 & \underline{0.2284} & \underline{0.2658} & 0.2363 & 0.2719 & 0.2285 & 0.2667 & 0.2443 & 0.2775 & 0.2338 & 0.2749 & 0.2416 & 0.2941 \\
ETTh1       & 0.4299 & 0.4429 & 0.4409 & 0.4516 & \textbf{0.4154} & \textbf{0.4346} & 0.4281 & 0.4434 & 0.4265 & \underline{0.4356} & 0.4485 & 0.4612 & 0.4600 & 0.4643 & \underline{0.4184} & 0.4386 \\
ETTh2       & \textbf{0.3532} & \textbf{0.3995} & 0.3686 & 0.4056 & \underline{0.3624} & 0.4052 & 0.3666 & \underline{0.4020} & 0.3742 & 0.4065 & 0.3846 & 0.4154 & 0.4036 & 0.4305 & 0.4995 & 0.4803 \\
ETTm1       & \textbf{0.3480} & \underline{0.3823} & 0.3790 & 0.3966 & \underline{0.3509} & 0.3835 & 0.3580 & 0.3869 & 0.3596 & 0.3899 & 0.3689 & 0.3989 & 0.3770 & 0.4048 & 0.3586 & \textbf{0.3816} \\
ETTm2       & \textbf{0.2554} & \textbf{0.3202} & 0.2649 & 0.3273 & \underline{0.2599} & 0.3236 & 0.2628 & \underline{0.3203} & 0.2663 & 0.3288 & 0.2717 & 0.3312 & 0.2798 & 0.3353 & 0.2768 & 0.3422 \\
Electricity & 0.1702 & 0.2674 & 0.1719 & 0.2720 & 0.1687 & 0.2660 & 0.1664 & \underline{0.2611} & 0.1664 & 0.2630 & \textbf{0.1640} & \textbf{0.2584} & \underline{0.1661} & 0.2662 & 0.1671 & 0.2675 \\
\hline
\end{tabular}}
\end{table*}

As shown in Table~\ref{tab:full_results_avg}, LLM-based approaches generally compare favorably with conventional time series forecasting models across most benchmark datasets. CASP-LLM attains the lowest average MSE on Weather, ETTh2, ETTm1, and ETTm2, and the lowest average MAE on Weather, ETTh2, and ETTm2; on ETTh1, $\text{S}^2$IP-LLM records the lowest averages.

Among the competing LLM-based methods, $\text{S}^2$IP-LLM is competitive on several datasets, indicating that semantic alignment between prompts and inputs is beneficial for long-term forecasting. CASP-LLM reports lower average errors than $\text{S}^2$IP-LLM on four of the six benchmarks, which we attribute to coverage-aware regularization reshaping the anchor embedding geometry; ETTh1, where $\text{S}^2$IP-LLM leads on the horizon average, is driven by horizons 192 and 720. On the Electricity dataset, iTransformer obtains the best overall performance, suggesting that strong inductive biases for periodic and highly regular variates remain effective in such domains. The averaged view is consistent with the per-horizon observations in the main paper that semantic alignment combined with coverage-aware regularization attains the lowest average error on the majority of these benchmarks.

%% file: 12_app_shortterm.tex
\section{Short-term Forecasting Results}
\label{app:shortterm}

We additionally evaluate CASP-LLM on the M4 short-term forecasting benchmark~\cite{makridakis2020m4}. The M4 dataset contains time series spanning a wide range of domains; following common practice, we evaluate on prediction horizons of 6 to 48 steps depending on the sampling frequency. We report symmetric Mean Absolute Percentage Error (sMAPE), Mean Absolute Scaled Error (MASE), and Overall Weighted Average (OWA).

Table~\ref{tab:short_full} reports the three principal frequency splits (Yearly, Quarterly, Monthly) together with the cross-frequency average, which aggregates all six M4 frequencies following the standard evaluation protocol. CASP-LLM achieves the best sMAPE and MASE on the Yearly split and the best MASE and OWA on the Quarterly split, while remaining competitive on the Monthly split where OFA attains the lowest sMAPE and MASE. In terms of the cross-frequency average, CASP-LLM obtains the best sMAPE and MASE, with $\text{S}^2$IP-LLM showing the strongest OWA. These results indicate that the coverage-aware retrieval mechanism transfers to the short-horizon regime in addition to the long-horizon setting reported in the main paper, with stable and competitive performance across temporal resolutions.

\begin{table*}[h!]
\centering
\caption{Short-term forecasting results on the M4 dataset (horizons 6--48). We report the three principal frequency splits (Yearly, Quarterly, Monthly); the Average follows the official M4 protocol and aggregates all six frequencies (Yearly through Hourly). Best in bold, second-best underlined.}
\label{tab:short_full}
\scriptsize
\begin{tabular}{c||c||c|c|c|c|c|c|c|c|c}
\hline
 &  & \textbf{Ours} & \textbf{Time-LLM} & \textbf{S$^2$IP-LLM} & \textbf{OFA} & \textbf{PatchTST} & \textbf{iTransformer} & \textbf{DLinear} & \textbf{Autoformer} & \textbf{FEDformer}\\
\hline
\midrule
\multirow{3}{*}{Yearly}
 & sMAPE & \textbf{13.280} & 13.721 & 13.899 & 14.368 & \underline{13.664} & 14.282 & 14.358 & 18.118 & 14.214 \\
 & MASE  & \textbf{3.003}  & \underline{3.036} & 3.093 & 3.397 & 3.086 & 3.202 & 3.128 & 4.131 & 3.166 \\
 & OWA   & \underline{0.806} & \textbf{0.802} & 0.814 & 0.867 & 0.806 & 0.840 & 0.833 & 1.074 & 0.833 \\
\hline
\multirow{3}{*}{Quarterly}
 & sMAPE & 10.509 & 10.861 & 10.545 & \textbf{10.444} & 10.916 & 10.740 & \underline{10.500} & 14.516 & 10.713 \\
 & MASE  & \textbf{1.204} & 1.293 & 1.239 & \underline{1.237} & 1.289 & 1.280 & 1.239 & 1.856 & 1.282 \\
 & OWA   & \textbf{0.912} & 0.965 & 0.931 & \underline{0.925} & 0.966 & 0.954 & 0.929 & 1.336 & 0.954 \\
\hline
\multirow{3}{*}{Monthly}
 & sMAPE & 13.543 & 13.445 & \underline{13.122} & \textbf{12.871} & 14.180 & 15.062 & 13.394 & 18.027 & 13.704 \\
 & MASE  & 0.998  & 1.027 & \underline{0.987} & \textbf{0.953} & 1.127 & 1.170 & 1.007 & 1.563 & 1.070 \\
 & OWA   & 0.962  & 0.949 & \underline{0.919} & \textbf{0.894} & 1.021 & 1.072 & 0.938 & 1.359 & 0.978 \\
\hline
\multirow{3}{*}{Average}
 & sMAPE & \textbf{12.218} & 12.522 & 12.314 & 12.256 & 12.873 & 13.359 & 12.507 & 16.636 & 12.666 \\
 & MASE  & \textbf{1.678} & 1.704 & \underline{1.684} & 1.714 & 1.757 & 1.786 & \underline{1.684} & 2.385 & 1.695 \\
 & OWA   & \underline{0.899} & 0.907 & \textbf{0.894} & 0.900 & 0.934 & 0.959 & 0.901 & 1.237 & 0.910 \\
\hline
\end{tabular}
\end{table*}

%% file: 13_app_fewshot.tex
\section{Few-Shot Forecasting}
\label{app:fewshot}

To assess data efficiency in low-resource regimes, we evaluate all models on 10\% of the original training data, keeping validation and test splits unchanged. The motivation is twofold: (i) coverage-aware retrieval, by encouraging broader use of the semantic anchor pool, may mitigate overfitting under limited training data, and (ii) low-data settings are common in real forecasting applications where labels are expensive or recent history is short.

Table~\ref{tab:fewshot} reports MSE and MAE under the 10\% training-data setting on the four ETT datasets. CASP-LLM yields the best results on the short-to-medium horizons (96, 192, 336) across all four datasets, with the largest relative gains on ETTm1 and ETTm2. The ordering at horizon 720 reverses: $\text{S}^2$IP-LLM and OFA are ahead on all four datasets, and DLinear is ahead on ETTm1 (0.411 against 0.487) and ETTm2 (0.316 against 0.366). The coverage term acts through a usage statistic accumulated across training batches, and this setting supplies the fewest such batches of any configuration we evaluate, so the statistic is least converged here (Section~\ref{sec:limitations}). That $\text{S}^2$IP-LLM, which shares the anchor architecture without the coverage term, is ahead at this horizon is consistent with a limit on the coverage statistic under few updates rather than with a limit on retrieval-based conditioning as such, though the few-shot runs are not paired at $\lambda_{\text{cov}}=0$ and so do not isolate that.

\begin{table*}[h!]
\centering
\scriptsize
\caption{Few-shot forecasting results (10\% of training data). Lower MSE/MAE is better. The best result in each row is in bold. Each cell is the average over three runs.}
\label{tab:fewshot}
\resizebox{\textwidth}{!}{%
\begin{tabular}{l|c||cc|cc|cc|cc|cc|cc|cc}
\toprule
 &  & \multicolumn{2}{c|}{\textbf{Ours}} & \multicolumn{2}{c|}{\textbf{S$^2$IP-LLM}} & \multicolumn{2}{c|}{\textbf{Time-LLM}} & \multicolumn{2}{c|}{\textbf{OFA}} & \multicolumn{2}{c|}{\textbf{iTransformer}} & \multicolumn{2}{c|}{\textbf{DLinear}} & \multicolumn{2}{c}{\textbf{PatchTST}} \\
\textbf{Dataset} & \textbf{H} & MSE & MAE & MSE & MAE & MSE & MAE & MSE & MAE & MSE & MAE & MSE & MAE & MSE & MAE \\
\midrule
\multirow{4}{*}{ETTh1}
 & 96  & \textbf{0.437} & \textbf{0.451} & 0.518 & 0.491 & 0.747 & 0.545 & 0.570 & 0.516 & 0.837 & 0.609 & 0.565 & 0.538 & 0.598 & 0.524 \\
 & 192 & \textbf{0.553} & \textbf{0.511} & 0.664 & 0.570 & 0.793 & 0.551 & 0.608 & 0.535 & 0.780 & 0.575 & 0.721 & 0.622 & 0.657 & 0.550 \\
 & 336 & \textbf{0.574} & \textbf{0.530} & 0.711 & 0.584 & 0.880 & 0.584 & 0.725 & 0.591 & 1.234 & 0.811 & 0.986 & 0.743 & 0.762 & 0.610 \\
 & 720 & 0.652 & 0.573 & 0.593 & 0.529 & 0.785 & 0.553 & \textbf{0.590} & \textbf{0.525} & 0.910 & 0.860 & 0.691 & 0.600 & 0.633 & 0.542 \\
\midrule
\multirow{4}{*}{ETTh2}
 & 96  & \textbf{0.354} & \textbf{0.371} & 0.401 & 0.423 & 0.430 & 0.438 & 0.402 & 0.411 & 0.470 & 0.474 & 0.569 & 0.519 & 0.403 & 0.414 \\
 & 192 & \textbf{0.382} & \textbf{0.413} & 0.442 & 0.450 & 0.449 & 0.458 & 0.406 & 0.433 & 0.489 & 0.485 & 0.671 & 0.572 & 0.426 & 0.441 \\
 & 336 & \textbf{0.401} & \textbf{0.434} & 0.480 & 0.486 & 0.485 & 0.490 & 0.449 & 0.464 & 0.593 & 0.538 & 0.824 & 0.648 & 0.477 & 0.480 \\
 & 720 & 0.453 & 0.471 & 0.419 & 0.439 & 0.424 & 0.441 & \textbf{0.397} & \textbf{0.421} & 0.489 & 0.483 & 0.605 & 0.538 & 0.415 & 0.431 \\
\midrule
\multirow{4}{*}{ETTm1}
 & 96  & \textbf{0.348} & \textbf{0.392} & 0.422 & 0.421 & 0.447 & 0.438 & 0.429 & 0.423 & 0.717 & 0.548 & 0.382 & 0.412 & 0.437 & 0.434 \\
 & 192 & \textbf{0.389} & \textbf{0.410} & 0.456 & 0.430 & 0.497 & 0.465 & 0.469 & 0.439 & 0.735 & 0.575 & 0.419 & 0.434 & 0.476 & 0.454 \\
 & 336 & \textbf{0.421} & \textbf{0.433} & 0.554 & 0.490 & 0.594 & 0.521 & 0.569 & 0.498 & 0.752 & 0.584 & 0.490 & 0.477 & 0.681 & 0.556 \\
 & 720 & 0.487 & 0.469 & 0.455 & 0.435 & 0.487 & 0.461 & 0.464 & 0.441 & 0.728 & 0.565 & \textbf{0.411} & \textbf{0.429} & 0.501 & 0.466 \\
\midrule
\multirow{4}{*}{ETTm2}
 & 96  & \textbf{0.189} & \textbf{0.281} & 0.246 & 0.313 & 0.260 & 0.317 & 0.251 & 0.309 & 0.274 & 0.338 & 0.278 & 0.345 & 0.252 & 0.317 \\
 & 192 & \textbf{0.240} & \textbf{0.320} & 0.301 & 0.340 & 0.312 & 0.349 & 0.307 & 0.346 & 0.361 & 0.394 & 0.338 & 0.385 & 0.306 & 0.353 \\
 & 336 & \textbf{0.284} & \textbf{0.342} & 0.400 & 0.403 & 0.424 & 0.416 & 0.426 & 0.417 & 0.467 & 0.442 & 0.436 & 0.440 & 0.433 & 0.427 \\
 & 720 & 0.366 & 0.392 & \textbf{0.284} & \textbf{0.332} & 0.305 & 0.344 & 0.293 & 0.335 & 0.336 & 0.373 & 0.316 & 0.368 & 0.296 & 0.343 \\
\bottomrule
\end{tabular}%
}
\end{table*}

%% file: 14_app_sensitivity.tex
\section{Hyperparameter Sensitivity}
\label{app:sensitivity}

This appendix studies the sensitivity of CASP-LLM to the coverage coefficient $\lambda_{\text{cov}}$, complementing the entropy and ablation results in the main paper. We analyze two effects: how $\lambda_{\text{cov}}$ shapes the empirical distribution of anchor selection (Section~\ref{app:sensitivity:concentration}), and how it reshapes the geometry of the learned anchor, time-series, and prompt embeddings (Section~\ref{app:sensitivity:pca}).

\subsection{Effect of Coverage Regularization on Anchor Selection Concentration}
\label{app:sensitivity:concentration}

\begin{figure*}[h!]
\centering
\includegraphics[width=\textwidth]{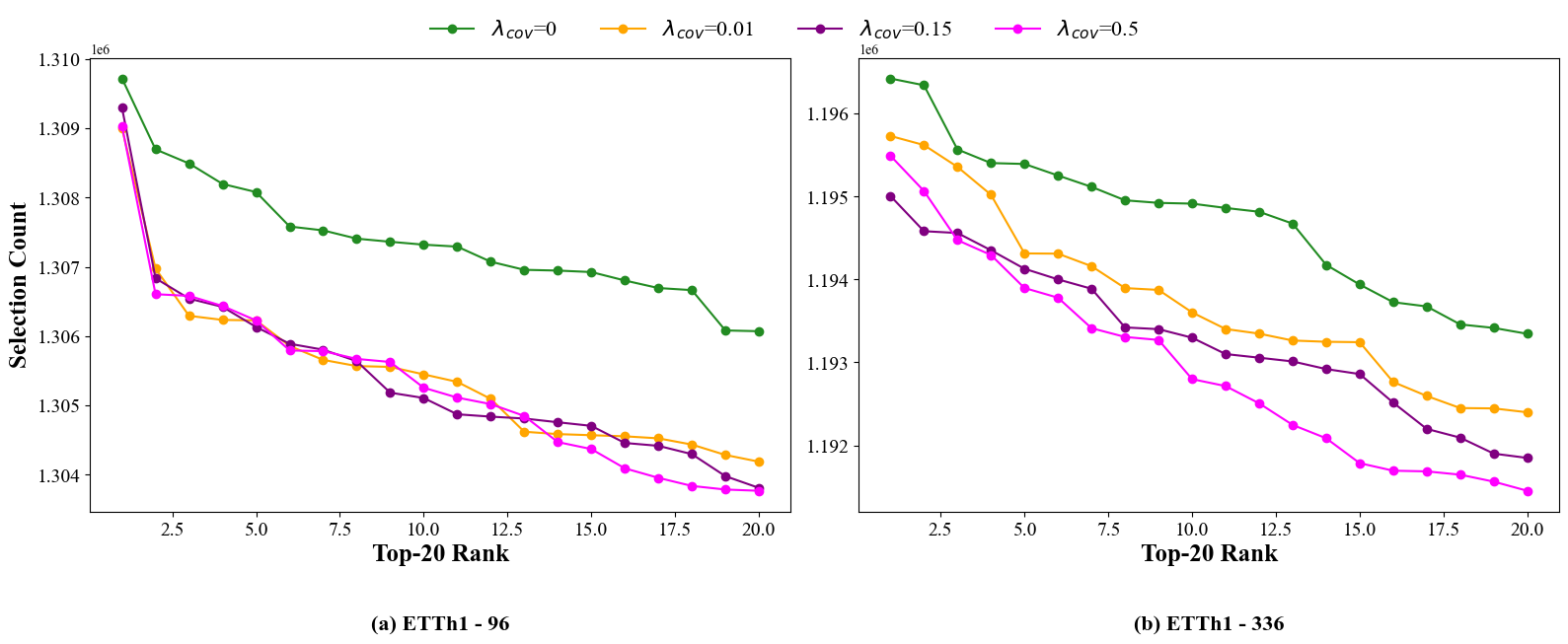}
\caption{Anchor selection concentration during training across $\lambda_{\text{cov}}$ on ETTh1-96 (left) and ETTh1-336 (right). Curves plot cumulative selection counts of the top-20 most frequently selected anchors.}
\label{fig:concentration_topk}
\end{figure*}

To understand how the coverage-aware selection mechanism affects the diversity of anchor usage, we plot the selection counts of the top-20 most frequently used anchors across values of the coverage coefficient $\lambda_{\text{cov}}$. Figure~\ref{fig:concentration_topk} reports the distributions for two representative forecasting horizons on ETTh1; this sweep used $\lambda_{\text{cov}}=0.15$ in place of the $0.1$ grid point of Table~\ref{tab:lambda}.

In the left subplot (ETTh1-96), the $\lambda_{\text{cov}}=0$ curve sits above the regularized curves at every rank and declines slowly across the top-20, so the leading anchors accumulate the most selections and the gap between adjacent ranks is small. As $\lambda_{\text{cov}}$ increases from 0.01 to 0.5, the curves shift downward and decline more steeply, so the leading anchors accumulate fewer selections relative to the rest. The spread between the highest and lowest entries within the top-20 is smallest at $\lambda_{\text{cov}}=0.5$. The absolute differences are small relative to the counts themselves; what the panel shows is the ordering of the four settings, not the size of the effect.

The right subplot (ETTh1-336) orders the four settings the same way, at slightly lower counts overall. The $\lambda_{\text{cov}}=0.01$ curve already sits below the unregularized one, and the $\lambda_{\text{cov}}=0.15$ and $\lambda_{\text{cov}}=0.5$ curves are close to each other in both panels, with flatter gaps between adjacent ranks than at $\lambda_{\text{cov}}=0$. These curves describe anchor usage as it accumulates during training. They are not a statement about the converged model: measured at inference on the test split, the usage entropy sits at its floor of $\ln K$ under every coverage weight and also without the term (Table~\ref{tab:lambda}), so the trained model concentrates on a small fixed set either way. The two views are consistent, and taken together they locate the mechanism in the training dynamics rather than in the final selection distribution.

\subsection{Coverage Weight Analysis via PCA}
\label{app:sensitivity:pca}

\begin{figure*}[h!]
\centering
\includegraphics[width=0.7\textwidth]{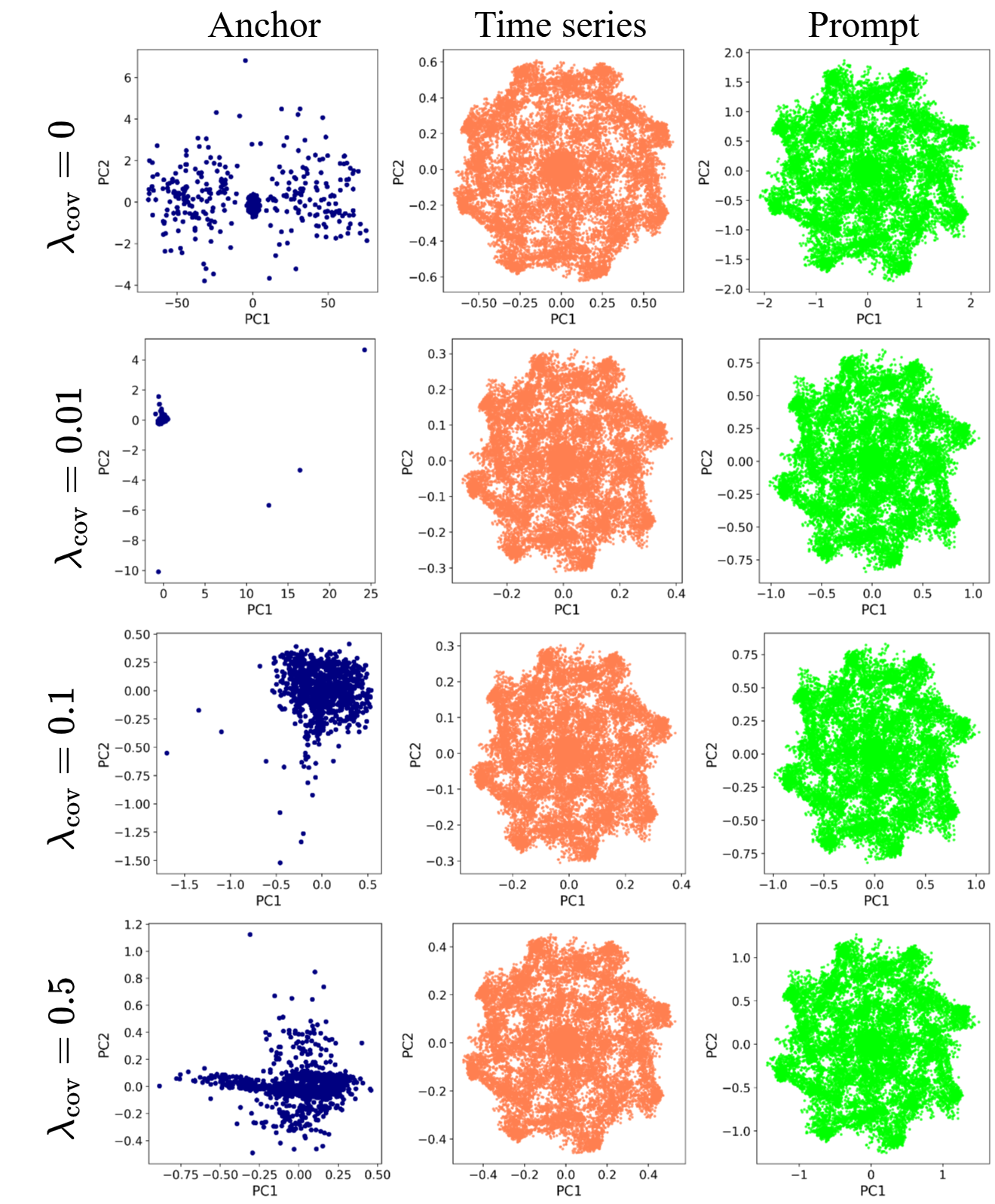}
\caption{PCA projections of anchor, time-series, and prompt embeddings under varying $\lambda_{\text{cov}}$.}
\label{fig:pca_projection}
\end{figure*}

\begin{table*}[h!]
\centering
\small
\renewcommand{\arraystretch}{1.2}
\setlength{\tabcolsep}{5pt}
\caption{PCA variance ratio of PC1 and PC2 across different $\lambda_{\text{cov}}$ values for anchor, time-series, and prompt embeddings.}
\label{tab:pca_embeddings}
\begin{tabular}{l|cc|cc|cc|cc}
\toprule
 & \multicolumn{2}{c|}{$\lambda_{\text{cov}}=0$} & \multicolumn{2}{c|}{$\lambda_{\text{cov}}=0.01$} & \multicolumn{2}{c|}{$\lambda_{\text{cov}}=0.1$} & \multicolumn{2}{c}{$\lambda_{\text{cov}}=0.5$} \\
\cline{2-9}
 & PC1 & PC2 & PC1 & PC2 & PC1 & PC2 & PC1 & PC2 \\
\midrule
Anchor embeddings      & 0.9850 & 0.0013 & 0.7185 & 0.1231 & 0.5583 & 0.4096 & 0.6226 & 0.3035 \\
\cline{2-9}
Total                  & \multicolumn{2}{c|}{\textbf{0.9862}} & \multicolumn{2}{c|}{\textbf{0.8416}} & \multicolumn{2}{c|}{\textbf{0.9679}} & \multicolumn{2}{c}{\textbf{0.9261}} \\
\midrule
Time-series embeddings & 0.3932 & 0.3338 & 0.3102 & 0.2504 & 0.3162 & 0.2557 & 0.3668 & 0.3024 \\
\cline{2-9}
Total                  & \multicolumn{2}{c|}{\textbf{0.7270}} & \multicolumn{2}{c|}{\textbf{0.5606}} & \multicolumn{2}{c|}{\textbf{0.5719}} & \multicolumn{2}{c}{\textbf{0.6692}} \\
\midrule
Prompt embeddings      & 0.4161 & 0.3434 & 0.3160 & 0.2514 & 0.3223 & 0.2560 & 0.3788 & 0.3063 \\
\cline{2-9}
Total                  & \multicolumn{2}{c|}{\textbf{0.7595}} & \multicolumn{2}{c|}{\textbf{0.5675}} & \multicolumn{2}{c|}{\textbf{0.5783}} & \multicolumn{2}{c}{\textbf{0.6851}} \\
\bottomrule
\end{tabular}
\end{table*}

\paragraph{Impact of the coverage coefficient on embedding geometry}
To assess the structural properties and information retention of the learned embeddings, we apply Principal Component Analysis (PCA) to project the high-dimensional embedding vectors into a two-dimensional space. We analyze the effect of $\lambda_{\text{cov}}$ on three types of embeddings, namely anchor embeddings, raw time-series embeddings, and prefix-prompted embeddings, to understand how coverage regularization influences the geometry and information distribution of their respective embedding spaces. The anchor analysis here covers the full pool of $V'$ anchors, whereas the cosine statistic in Table~\ref{tab:lambda} is computed over the selected subset; both collapse at $\lambda_{\text{cov}}=0$. For each setting of $\lambda_{\text{cov}} \in \{0, 0.01, 0.1, 0.5\}$, we project the embeddings onto their top two principal components and report the corresponding explained variance.

Formally, let $\mathbf{X} \in \mathbb{R}^{n \times d}$ denote a matrix of $n$ $d$-dimensional embedding vectors. We first compute the covariance matrix
\begin{equation*}
\mathbf{\Sigma} = \frac{1}{n-1} \mathbf{X}^\top \mathbf{X},
\end{equation*}
where $d$ is the embedding dimension and $n$ is the number of samples (anchors, time-series instances, or prompts). Eigendecomposition of $\mathbf{\Sigma}$ yields eigenvalues $\sigma_1^2, \sigma_2^2, \ldots, \sigma_d^2$, each representing the variance explained by the corresponding principal component. Let $\sigma_{\text{PC1}}^2$ and $\sigma_{\text{PC2}}^2$ denote the eigenvalues of the first and second principal components. The total explained variance ratio of the first two components is
\begin{equation*}
\text{Explained Variance Ratio} = \frac{\sigma_{\text{PC1}}^2 + \sigma_{\text{PC2}}^2}{\sum_{j=1}^{d} \sigma_j^2}.
\end{equation*}

As shown in Table~\ref{tab:pca_embeddings} and Figure~\ref{fig:pca_projection}, at $\lambda_{\text{cov}}=0$ the anchor embeddings exhibit a highly anisotropic distribution, with the majority of variance concentrated along PC1 ($\sigma_{\text{PC1}}^2 = 0.985$) and only $\sigma_{\text{PC2}}^2 = 0.0013$ on PC2, giving a total explained variance of 0.9862. This indicates a near-collapsed anchor space in which semantic anchors align along a single dominant direction. As $\lambda_{\text{cov}}$ increases to 0.1, PC1 and PC2 explain $\sigma_{\text{PC1}}^2 = 0.5583$ and $\sigma_{\text{PC2}}^2 = 0.4096$ of the variance respectively (total 0.9679), reflecting a more isotropic and balanced embedding structure. At $\lambda_{\text{cov}}=0.5$, the total explained variance remains relatively high at 0.9261, showing that stronger coverage regularization maintains information retention while promoting geometric dispersion.

In contrast, time-series embeddings display a stable variance allocation across all $\lambda_{\text{cov}}$ values. At $\lambda_{\text{cov}}=0$, PC1 and PC2 explain $\sigma_{\text{PC1}}^2 = 0.3932$ and $\sigma_{\text{PC2}}^2 = 0.3338$ of the variance (total 0.7270), and this shifts only mildly to a total of 0.6692 at $\lambda_{\text{cov}}=0.5$. This consistency suggests that raw time-series embeddings inherently reflect temporal structure and are largely insensitive to anchor-side regularization.

Prompted embeddings exhibit intermediate sensitivity. At $\lambda_{\text{cov}}=0$, PC1 and PC2 account for variance ratios of $0.4161$ and $0.3434$ (total 0.7595). As $\lambda_{\text{cov}}$ increases to 0.01 and 0.1, the total drops to 0.5675 and 0.5783, indicating that moderate regularization redistributes variance across more directions. The total then recovers to 0.6851 at $\lambda_{\text{cov}}=0.5$, suggesting that strong coverage constraints promote diversity while partially restoring an aligned structure. Across all values of $\lambda_{\text{cov}}$, prompted embeddings consistently retain more variance in the top two components than raw time-series embeddings.

In summary, the PCA analysis shows that anchor and prompted embeddings are sensitive to the coverage coefficient $\lambda_{\text{cov}}$, with moderate to strong values (0.1 or 0.5) effectively enhancing isotropy and mitigating anchor collapse. Time-series embeddings maintain a stable geometric structure regardless of coverage regularization, highlighting their robustness and independence from anchor selection dynamics.

%% file: 12_app_significance.tex
\section{Significance Testing}
\label{app:significance}

\subsection{Protocol}
\label{app:significance:protocol}

Every test below compares CASP-LLM against the $\lambda_{\text{cov}}=0$ configuration, which is the same model with the coverage term removed. The two conditions share code, data, fixed seeds, and tuned hyperparameters, so a paired test isolates the effect of the coverage term rather than a difference in implementation or tuning. We use the two-sided Wilcoxon signed-rank test throughout and report the test statistic $W$ alongside the $p$-value. Standard deviations are sample standard deviations. Main-table cells use three fixed seeds; the extended tests on the primary cells add five more, for eight paired seeds.

\subsection{Paired Cell-Level Comparison}
\label{app:significance:paired}

Table~\ref{tab:paired_cov0} reports the paired comparison for all 24 dataset--horizon cells. Coverage improves 14 of 24 cells. Over the 20 non-Electricity cells it improves 12 and the cross-cell signed-rank test over per-cell mean deltas gives $W=74$, $p=0.2611$. Including the four Electricity cells gives 14 of 24, $W=116$, $p=0.3449$. Neither cross-cell test reaches significance at $\alpha=0.05$.

\begin{table}[h!]
\centering
\scriptsize
\setlength{\tabcolsep}{3pt}
\resizebox{\columnwidth}{!}{%
\begin{tabular}{l|c||c|c|c|c}
\toprule
Dataset & $H$ & CASP-LLM & $\lambda_{\text{cov}}=0$ & $\Delta$ & $\Delta$\,\% \\
\midrule
\multirow{4}{*}{ETTh1} & 96 & $0.3714\pm0.0036$ & $0.3752\pm0.0067$ & $-0.0038$ & $-1.0$ \\
 & 192 & $0.4161\pm0.0139$ & $0.3987\pm0.0087$ & $+0.0174$ & $+4.4$ \\
 & 336 & $0.4263\pm0.0183$ & $0.4451\pm0.0025$ & $-0.0189$ & $-4.2$ \\
 & 720 & $0.5057\pm0.0296$ & $0.5063\pm0.0495$ & $-0.0007$ & $-0.1$ \\
\midrule
\multirow{4}{*}{ETTh2} & 96 & $0.2846\pm0.0040$ & $0.2826\pm0.0023$ & $+0.0019$ & $+0.7$ \\
 & 192 & $0.3474\pm0.0011$ & $0.3490\pm0.0033$ & $-0.0016$ & $-0.4$ \\
 & 336 & $0.3607\pm0.0051$ & $0.3606\pm0.0018$ & $+0.0000$ & $+0.0$ \\
 & 720 & $0.4203\pm0.0066$ & $0.4245\pm0.0024$ & $-0.0042$ & $-1.0$ \\
\midrule
\multirow{4}{*}{ETTm1} & 96 & $0.2965\pm0.0044$ & $0.2937\pm0.0032$ & $+0.0028$ & $+0.9$ \\
 & 192 & $0.3285\pm0.0023$ & $0.3313\pm0.0048$ & $-0.0027$ & $-0.8$ \\
 & 336 & $0.3557\pm0.0026$ & $0.3577\pm0.0009$ & $-0.0020$ & $-0.6$ \\
 & 720 & $0.4113\pm0.0033$ & $0.4128\pm0.0054$ & $-0.0015$ & $-0.4$ \\
\midrule
\multirow{4}{*}{ETTm2} & 96 & $0.1651\pm0.0024$ & $0.1655\pm0.0019$ & $-0.0004$ & $-0.3$ \\
 & 192 & $0.2255\pm0.0040$ & $0.2210\pm0.0032$ & $+0.0045$ & $+2.0$ \\
 & 336 & $0.2760\pm0.0046$ & $0.2762\pm0.0061$ & $-0.0002$ & $-0.1$ \\
 & 720 & $0.3549\pm0.0027$ & $0.3564\pm0.0059$ & $-0.0015$ & $-0.4$ \\
\midrule
\multirow{4}{*}{Weather} & 96 & $0.1483\pm0.0013$ & $0.1481\pm0.0007$ & $+0.0002$ & $+0.1$ \\
 & 192 & $0.1935\pm0.0007$ & $0.1947\pm0.0016$ & $-0.0012$ & $-0.6$ \\
 & 336 & $0.2445\pm0.0025$ & $0.2443\pm0.0023$ & $+0.0002$ & $+0.1$ \\
 & 720 & $0.3181\pm0.0026$ & $0.3180\pm0.0025$ & $+0.0002$ & $+0.1$ \\
\midrule
\multirow{4}{*}{Electricity} & 96 & $0.1342\pm0.0006$ & $0.1340\pm0.0011$ & $+0.0002$ & $+0.2$ \\
 & 192 & $0.1511\pm0.0002$ & $0.1512\pm0.0003$ & $-0.0001$ & $-0.1$ \\
 & 336 & $0.1736\pm0.0055$ & $0.1739\pm0.0043$ & $-0.0004$ & $-0.2$ \\
 & 720 & $0.2220\pm0.0062$ & $0.2190\pm0.0054$ & $+0.0030$ & $+1.4$ \\
\bottomrule
\end{tabular}}
\caption{Paired comparison against $\lambda_{\text{cov}}=0$, three seeds per cell. Negative $\Delta$ favors CASP-LLM.}
\label{tab:paired_cov0}
\end{table}

Per-cell margins are small relative to seed variability. The largest gain is ETTh1-336 at $-4.2\%$ and the largest loss is ETTh1-192 at $+4.4\%$; outside ETTh1, only ETTm2-192 and Electricity-720 move by more than $1\%$ in either direction. On Electricity the four cells sum to no measurable effect, and the longest horizon moves against coverage by $+1.4\%$.

\subsection{Extended Paired Tests on the Primary Cells}
\label{app:significance:eightseed}

Three seeds resolve cell-level differences of this size poorly, so we extend the two ETTh1 and two ETTm1 cells used for the mechanism analyses to eight paired seeds. Results are in Table~\ref{tab:eightseed}. ETTh1-96 improves in seven of eight seeds and is significant at $\alpha=0.05$. ETTh1-336 improves in six of eight but does not reach significance. Neither ETTm1 cell reaches significance, which is the expected outcome on a dataset where the per-cell margin in Table~\ref{tab:paired_cov0} is under $1\%$. Means in Table~\ref{tab:eightseed} are over eight seeds and therefore differ from the three-seed means in Table~\ref{tab:paired_cov0}.

\begin{table}[h!]
\centering
\scriptsize
\setlength{\tabcolsep}{4pt}
\resizebox{\columnwidth}{!}{%
\begin{tabular}{l||c|c|c|c|c}
\toprule
Cell & CASP-LLM & $\lambda_{\text{cov}}=0$ & Improved & $W$ & $p$ \\
\midrule
ETTh1-96  & $0.3702\pm0.0026$ & $0.3726\pm0.0050$ & 7/8 & 3.0  & $\mathbf{0.039}$ \\
ETTh1-336 & $0.4294\pm0.0114$ & $0.4369\pm0.0086$ & 6/8 & 7.0  & 0.148 \\
ETTm1-96  & $0.2944\pm0.0033$ & $0.2936\pm0.0030$ & 5/8 & 16.0 & 0.844 \\
ETTm1-336 & $0.3566\pm0.0021$ & $0.3577\pm0.0013$ & 6/8 & 5.0  & 0.078 \\
\bottomrule
\end{tabular}}
\caption{Eight-seed paired tests on the primary cells. Bold marks significance at $\alpha=0.05$.}
\label{tab:eightseed}
\end{table}

\subsection{Stratification by Input Difficulty}
\label{app:significance:strata}

We label each test window by the residual energy of its lookback under the same classical decomposition the model applies, taking the mean absolute residual after removing a centered moving-average trend and a period-averaged seasonal component, averaged over channels. This is the most direct available proxy for the failure mode the coverage term targets, and we fix the labels before inspecting per-stratum errors. Over 64 (cell, seed) pairs, spanning the 20 non-Electricity cells at three seeds with ETTh1-96 extended to seven, the hardest decile changes by $-0.64\%$ with 40 of 64 pairs improving ($W=782.0$, $p=0.085$), while the easiest 30\% of windows moves by $-0.02\%$ with 31 of 64 improving ($W=1005.5$, $p=0.818$). The contrast between the two strata is $-0.63\%$. The hard-stratum test is marginal and does not reach $\alpha=0.05$; what it supports is that the gains concentrate on hard inputs and that no accuracy on easy inputs is traded away for them, not that the hard-stratum effect is itself established.

\subsection{Two Further Stratification Axes}
\label{app:significance:axes}

Input residual is one of three axes along which the coverage term might concentrate. The other two are windows that contain a trend change and windows dominated by low-frequency content, and we specify both here; the corresponding analyses are not included in this paper.

For the trend-change axis, a window is labeled by applying changepoint detection to its lookback trend, using PELT~\cite{killick2012pelt} with an $\ell_2$ cost over the centered moving-average trend of the lookback only, so no label depends on the forecast target. Labels are fixed before any per-stratum error is inspected, and results are read across a sweep of the PELT penalty rather than at a single value, since the segment count is sensitive to it. The trend-strength statistic of Wang et al.~\cite{wang2006characteristic}, the share of variance in the detrended series attributable to the trend component, serves as a second labeler, so that a stratification effect is not an artifact of one detector. A multivariate treatment is needed for the benchmarks used here: running the detector on the channel average tests a different hypothesis from a per-channel detector combined by a union rule, and the two need not agree on which windows contain a shift.

For the low-frequency axis, a window would be labeled by the share of its lookback spectrum below a cutoff, with the cutoff set from the sampling frequency rather than fitted. A regime dominated by low-frequency content is one where a small number of anchors can cover the input well, so it is the setting in which coverage would be expected to matter least, and it is the natural negative control for the input-residual result in Section~\ref{app:significance:strata}.

\subsection{Usage-Balancing Objectives}
\label{app:significance:balancing}

Table~\ref{tab:balancing} substitutes three usage-balancing objectives for the coverage term on the primary cells, each with the same soft-similarity signal and the same injection point and acting as the sole usage-shaping term: entropy regularization of the usage distribution, KL divergence toward uniform usage, and the load-balancing loss of Switch Transformers~\cite{fedus2022switch}. KL-to-uniform is gradient-equivalent to the entropy term on the usage simplex, so the entropy row also covers it. The objectives were not tuned per cell, matching the MMR protocol in Section~\ref{sec:discussion:connections}. Entropy regularization leaves the selection distribution where it was: every cell sits at $\ln K$ to four decimals, with exactly $K$ distinct anchors selected across the test split and no variance across seeds. Load balancing is the only objective that moves usage, and it moves it far off the floor, selecting hundreds of distinct anchors across the test split on three of the four cells (17 to 25 on ETTh1-336) against budgets of $K=8$ and $K=4$; it has the highest error at both ETTh1 horizons, while on ETTm1 the four conditions lie within seed spread of one another. Coverage stays at or near the floor and alone lowers error at both ETTh1 horizons. Usage dispersion and geometric dispersion also come apart here: under load balancing the mean pairwise cosine among selected keys, the Table~\ref{tab:lambda} statistic, stays between 0.73 and 0.91 while usage entropy is far above the floor, with ETTm1-96 selecting hundreds of anchors at cosine 0.91, the same separation Table~\ref{tab:lambda} shows along the $\lambda_{\text{cov}}$ axis. Neither spreading the usage histogram nor holding it fixed lowers error, which is consistent with the key-geometry account in Section~\ref{sec:exp:ablation}.

\begin{table}[h!]
\centering
\caption{Usage-balancing objectives substituted for the coverage term on the primary cells, three paired seeds. Entropy is hard-assignment usage entropy on the test split, with floor $\ln K$: $2.079$ for ETTh1 ($K=8$) and $1.386$ for ETTm1 ($K=4$).}
\scriptsize
\setlength{\tabcolsep}{3pt}
\resizebox{\columnwidth}{!}{%
\begin{tabular}{c|c|l||c|c|c|c}
\toprule
 &  & & $\lambda_{\text{cov}}=0$ & Entropy reg. & Load balance & Coverage \\
\midrule
\multirow{4}{*}{ETTh1} & \multirow{2}{*}{96} & Entropy & $2.0794$ & $2.0794$ & $4.9442$\stdv{0.5136} & $2.0991$\stdv{0.0341} \\
  & & MSE & $0.3752$\stdv{0.0067} & $0.3767$\stdv{0.0079} & $0.3876$\stdv{0.0064} & $\mathbf{0.3714}$\stdv{0.0036} \\
\cline{2-7}
  & \multirow{2}{*}{336} & Entropy & $2.0794$ & $2.0794$ & $2.4155$\stdv{0.1498} & $2.0794$ \\
  & & MSE & $0.4451$\stdv{0.0025} & $0.4366$\stdv{0.0192} & $0.4499$\stdv{0.0038} & $\mathbf{0.4263}$\stdv{0.0183} \\
\midrule
\multirow{4}{*}{ETTm1} & \multirow{2}{*}{96} & Entropy & $1.3863$ & $1.3863$ & $5.4030$\stdv{0.2108} & $1.3975$\stdv{0.0194} \\
  & & MSE & $0.2937$\stdv{0.0032} & $0.2943$\stdv{0.0027} & $\mathbf{0.2927}$\stdv{0.0027} & $0.2965$\stdv{0.0044} \\
\cline{2-7}
  & \multirow{2}{*}{336} & Entropy & $1.3863$ & $1.3863$ & $5.3758$\stdv{0.1513} & $1.3863$ \\
  & & MSE & $0.3577$\stdv{0.0009} & $0.3578$\stdv{0.0008} & $0.3577$\stdv{0.0059} & $\mathbf{0.3557}$\stdv{0.0026} \\
\bottomrule
\end{tabular}}
\label{tab:balancing}
\end{table}

\subsection{What the Tests Establish}
\label{app:significance:summary}

Read together, the tests place the coverage term as follows. Its effect is statistically resolved on one cell, ETTh1-96, at eight paired seeds. Elsewhere it is directional: 14 of 24 cells improve and both cross-cell tests are non-significant, so the aggregate evidence is consistent with a small positive effect but does not establish one. The effect concentrates on high-residual inputs, marginally. Two regimes fall outside it: Electricity, where the four cells show no measurable effect and the longest horizon moves against coverage, and the few-shot long-horizon setting discussed in Section~\ref{sec:limitations}, where CASP-LLM trails $\text{S}^2$IP-LLM and OFA on every ETT dataset and DLinear on ETTm1 and ETTm2. The mechanism analyses in Section~\ref{sec:exp:ablation} use ETTh1-96 as the primary cell because it is the one cell where the paired test resolves.

%% file: 15_app_qualitative.tex
\section{Embedding Alignment Visualization}
\label{app:qualitative}

To complement the t-SNE visualization in the main paper (Figure~\ref{fig:tsne_cov_ablation}), we provide additional embedding-level visualizations comparing raw time-series embeddings, prefix-prompted embeddings, and their pairwise similarity.

\begin{figure}[h!]
    \centering
    \begin{subfigure}[t]{0.32\columnwidth}
        \centering
        \includegraphics[width=\linewidth]{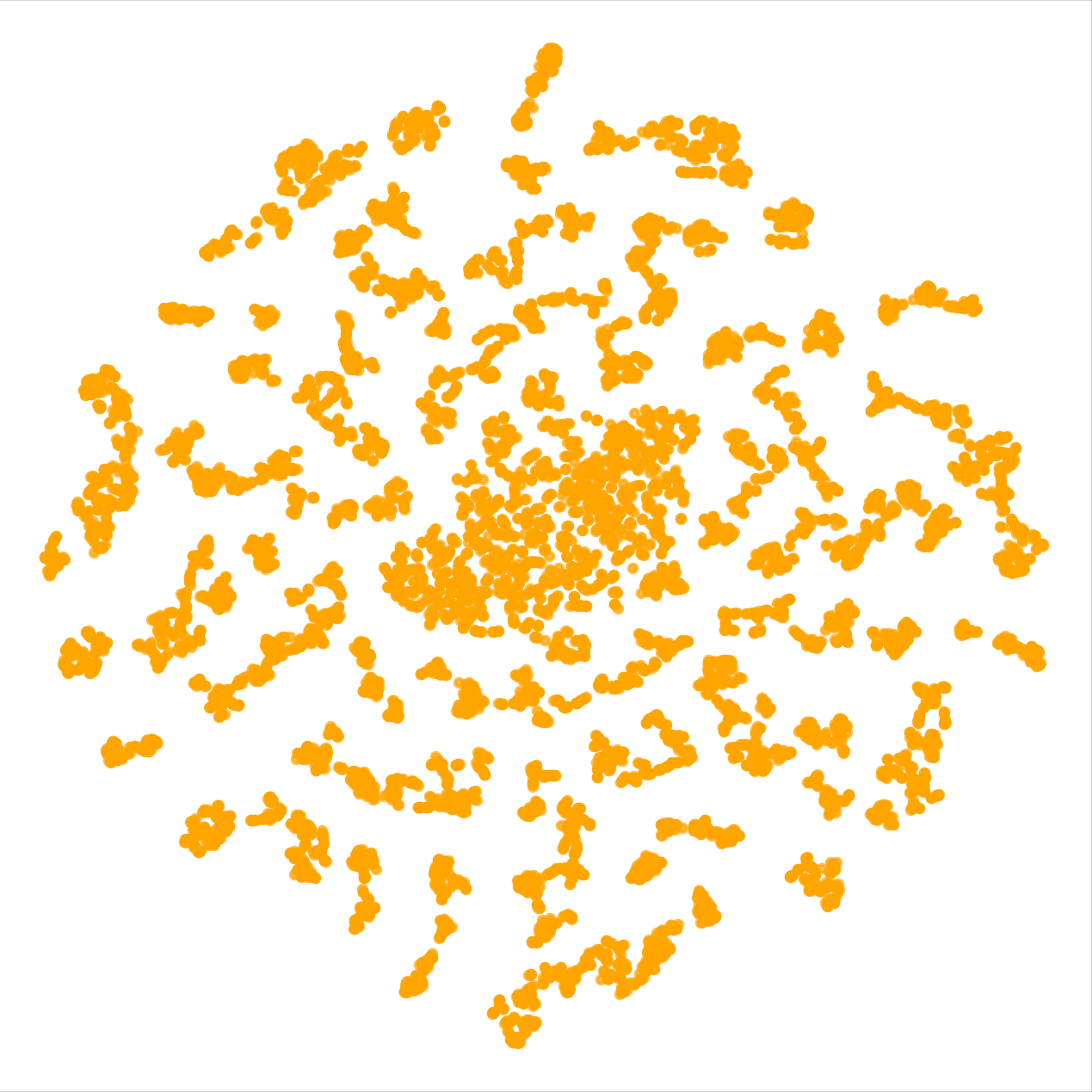}
        \caption{ }
        \label{fig:tsem}
    \end{subfigure}
    \hfill
    \begin{subfigure}[t]{0.32\columnwidth}
        \centering
        \includegraphics[width=\linewidth]{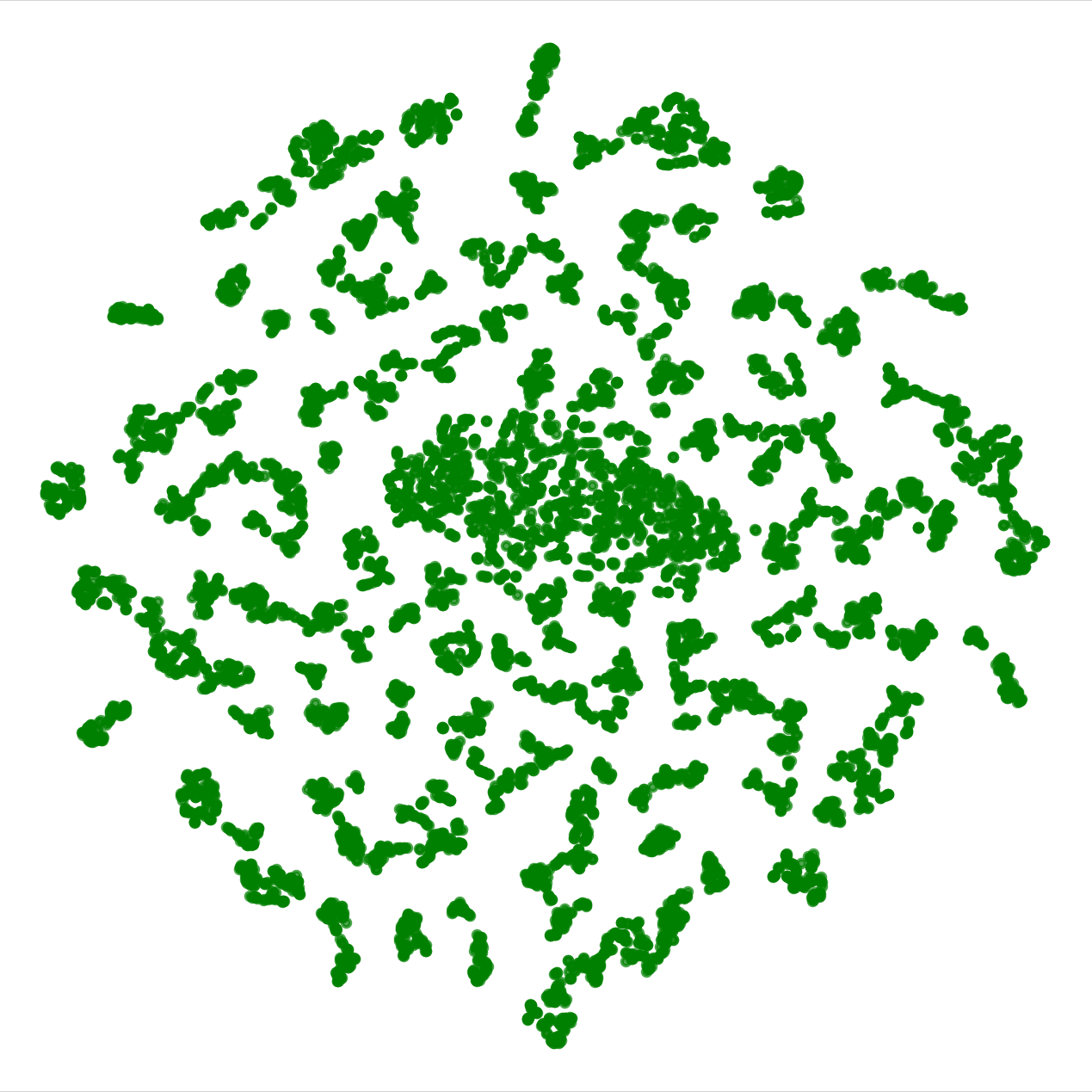}
        \caption{ }
        \label{fig:psem}
    \end{subfigure}
    \hfill
    \begin{subfigure}[t]{0.32\columnwidth}
        \centering
        \includegraphics[width=\linewidth]{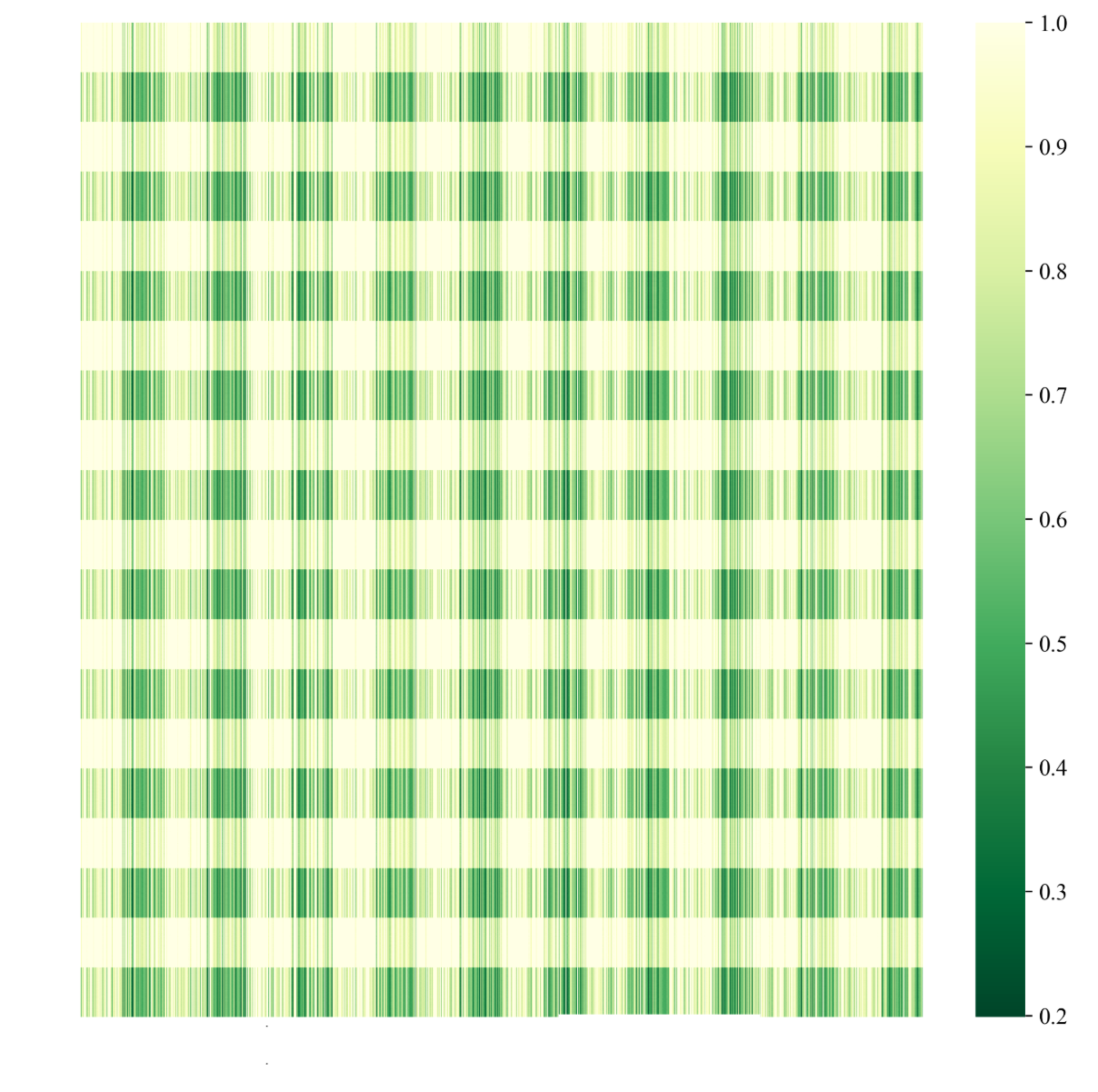}
        \caption{ }
        \label{fig:map}
    \end{subfigure}
\caption{Embedding visualization and alignment. (a) time series embeddings, (b) prefix-prompted embeddings, and (c) their cosine similarity map highlight structural consistency.}
\label{fig:simi}
\end{figure}

Figure~\ref{fig:simi} offers a detailed view of how semantic prompting influences structural alignment. Figure~\ref{fig:tsem} presents the t-SNE projection of raw time series embeddings, while Figure~\ref{fig:psem} shows the corresponding prefix embeddings. Both exhibit similar radial structures, suggesting that prefix-based representations preserve the global characteristics of the original input, with prefix embeddings forming more compact and coherent clusters that reflect enhanced semantic alignment. Figure~\ref{fig:map} visualizes the similarity between time-series and prefix embeddings, where the map shows a periodic band structure along both axes rather than a single diagonal: the same subsets of windows recur at regular intervals in the time-series and prefix representations, and similarity stays high across the map. Together, these visualizations indicate that our framework reorganizes the latent space along semantically meaningful dimensions while preserving structural fidelity.